%% file: main.tex
\documentclass[runningheads]{llncs}

\usepackage[mobile]{arxiv}

\usepackage{eccvabbrv}

\usepackage{amsmath,amssymb}
\usepackage{bm}
\usepackage{graphicx}
\usepackage{booktabs}
\usepackage{multirow}
\usepackage{array}
\usepackage{makecell}
\usepackage{tabularx}
\usepackage{colortbl}
\usepackage{float}
\usepackage[table,dvipsnames]{xcolor}
\usepackage{tikz}
\usetikzlibrary{fadings,patterns}
\usepackage{pgfplots}
\pgfplotsset{compat=1.17}
\usepackage[most]{tcolorbox}
\usepackage{enumitem}
\usepackage[normalem]{ulem}

\renewcommand{\lablogotextfont}{\vspace{2.7mm}\scriptsize\color{gray}} 
\lablogo[1.7cm]{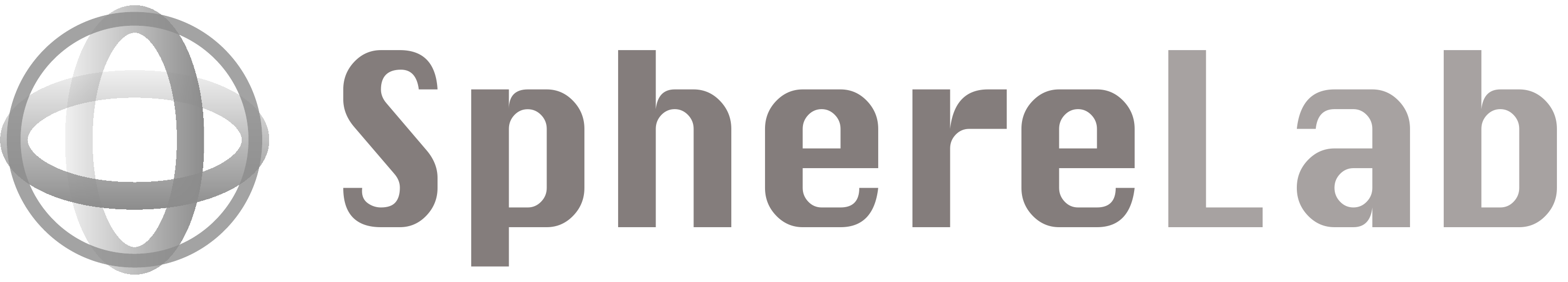}      %
\lablogotext{ECCV 2026} 

\usepackage[pagebackref=true,breaklinks=true,colorlinks=true,bookmarks=false]{hyperref}

\definecolor{deepred}{HTML}{940000}
\hypersetup{linkcolor=deepred}
\hypersetup{urlcolor  = [rgb]{0.4,0.15,0.95}}
\hypersetup{citecolor=[rgb]{0.4,0.15,0.95}}

\usepackage{pifont}
\usepackage{tocloft}
\usepackage[toc,page,header]{appendix}
\usepackage{adjustbox}

\definecolor{lightorange}{rgb}{1.0,0.85,0.7}
\definecolor{CadetBlue}{RGB}{95,158,160}
\definecolor{myblue}{RGB}{50, 115, 180}
\definecolor{mygray}{RGB}{190, 190, 190}
\definecolor{DarkViolet}{HTML}{9400D3}
\definecolor{MediumSeaGreen}{HTML}{3CB371}
\definecolor{DeepPink}{HTML}{FF1493}
\definecolor{DarkCyan}{HTML}{008B8B}

\newcommand{\blueboxtext}[1]{%
    \tikz[baseline=(X.base)]\node[draw=blue, thick, rounded corners=2pt, inner sep=2pt, text=blue] (X) {#1};%
}

\newcommand{\red}[1]{\textcolor{red}{#1}}
\newcommand{\blue}[1]{\textcolor{blue}{#1}}

\newcommand{\agentlogo}{\raisebox{-.5em}[0pt][0pt]{\includegraphics[height=1.7em]{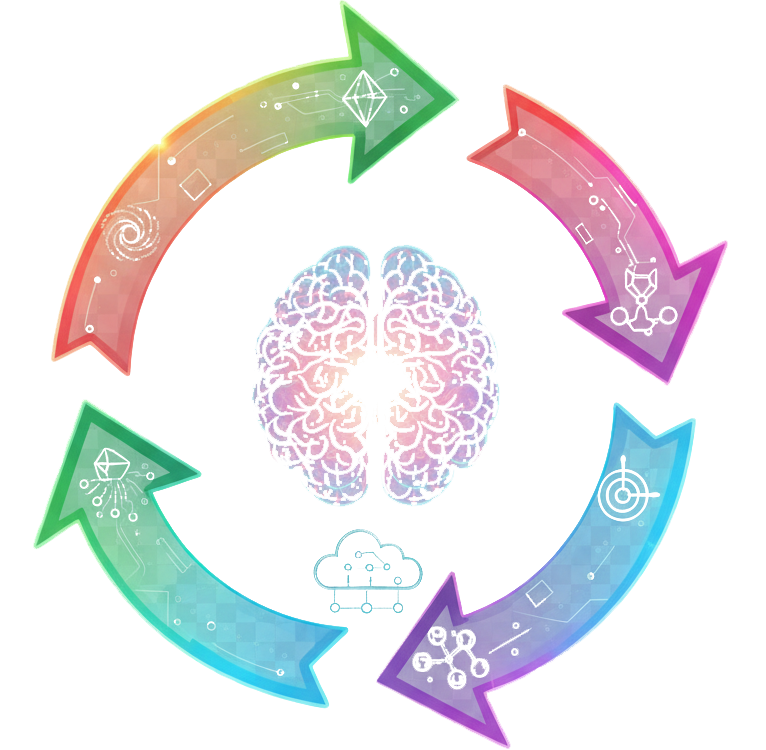}}}
\newcommand{\symbomni}{%
    \textcolor{orange!60!black}{S}%
    \textcolor{red!55!black}{y}%
    \textcolor{magenta!60!black}{m}%
    \textcolor{purple!60!black}{b}%
    \textcolor{blue!65!black}{O}%
    \textcolor{cyan!60!black}{m}%
    \textcolor{green!55!black}{n}%
    \textcolor{yellow!65!black}{i}%
}

\begin{document}
 
\title{\agentlogo\symbomni: Evolving Agentic Omni Models\\ via Symbolic Concept Learning}
\titlerunning{SymbOmni: Evolving Agentic Omni Models via Symbolic Concept Learning}

\author{\footnotesize Jinxiu Liu\inst{1,2}\textsuperscript{*} \and Jianru Li\inst{1}\textsuperscript{*} \and Tanqing Kuang\inst{1}\textsuperscript{*} \and Xuanming Liu\inst{2} \and Kangfu Mei\inst{3} \and
Yandong Wen\inst{2}\textsuperscript{\textdagger} \and Weiyang Liu\inst{4,5}}
\authorrunning{J. Liu, J. Li, T. Kuang, X. Liu, K. Mei, Y. Wen, W. Liu}
\institute{\scriptsize $^{1}$South China University of Technology~~~$^{2}$Westlake University~~~$^{3}$Johns Hopkins University\\[0.5mm] 
\scriptsize $^{4}$The Chinese University of Hong Kong~~~$^{5}$Shenzhen Loop Area Institute\\[3.5mm]
{\footnotesize\tt\href{http://spherelab.ai/symbomni}{\textbf{spherelab.ai/symbomni}}}
\vspace{-3mm}}

\maketitle

\begingroup
\renewcommand{\thefootnote}{}
\footnotetext{\scriptsize\textsuperscript{*}Equal contribution; authors may list themselves as first author. \textsuperscript{\textdagger}Corresponding author}
\endgroup

\begin{figure}[t]
\setlength{\abovecaptionskip}{4pt}
  \setlength{\belowcaptionskip}{-3pt}
    \centering
    \includegraphics[width=\textwidth]{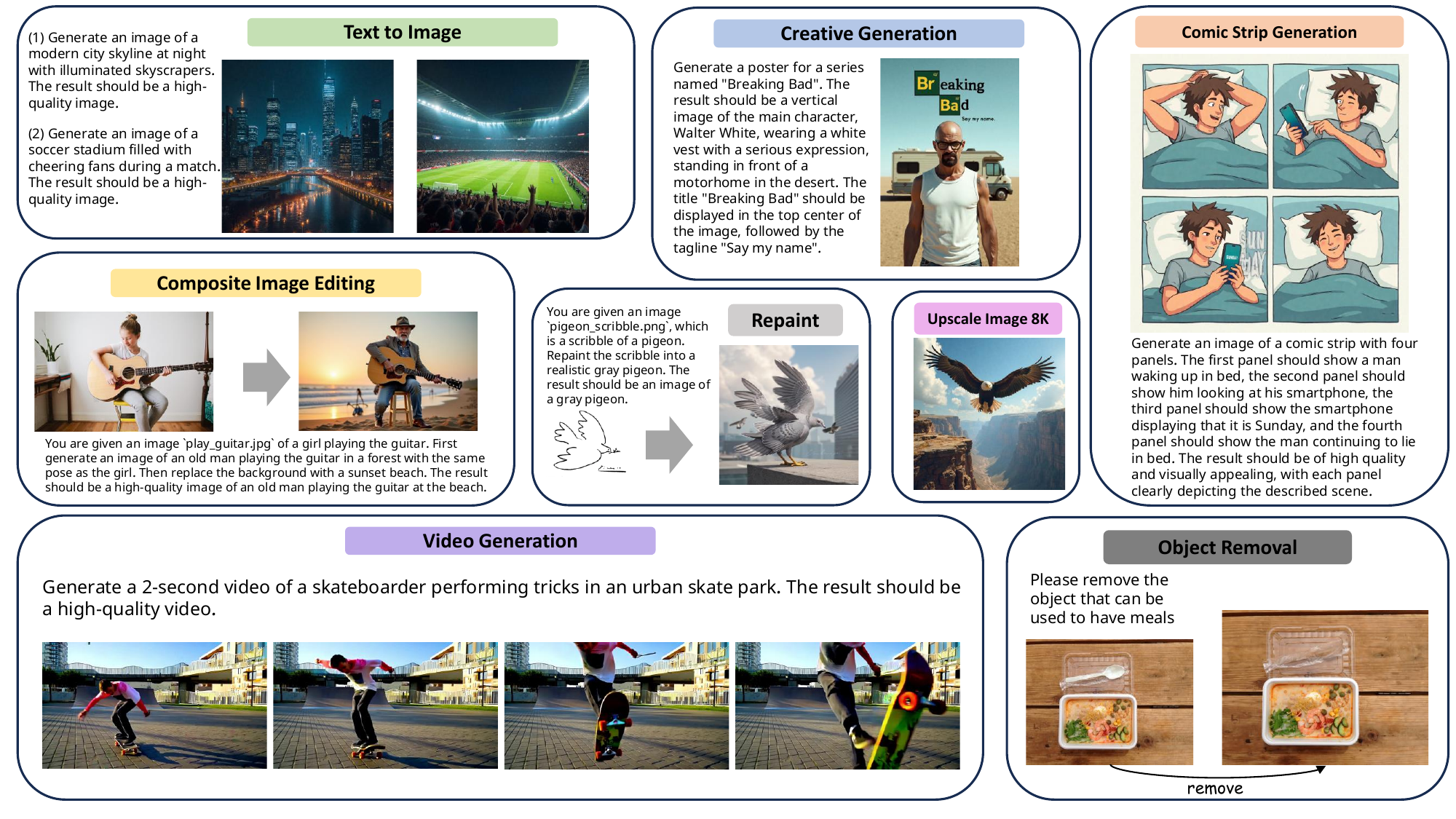}
    \caption{
Overview of the proposed SymbOmni framework. The model achieves versatile visual generation (text-to-image, creative content, image editing, video synthesis) through a unified approach for cumulative symbolic concept learning.
} \label{I2V-Z}
\end{figure}

\input{sec/0_abstract}
\input{sec/1_intro}

\input{sec/2_formatting}

\input{sec/3_finalcopy}

{
    \small
    \bibliographystyle{splncs04}
    \bibliography{main}
}

\newpage
\pagestyle{empty}
\begingroup
\makeatletter
\def\l@title#1#2{}%
\def\l@author#1#2{}%
\def\authcount#1{}%
\begin{center}
    {\Large\bfseries Appendix}
\end{center}
\vspace{6mm}
{\large\bfseries Table of Contents}\par
\vspace{1.5mm}
\hrule height 0.9pt
\vspace{1mm}
\renewcommand\@dotsep{10000}%
\def\l@section#1#2{\addvspace{11\p@}{\bfseries\@dottedtocline{1}{1.8em}{2em}{#1}{#2}}}%
\@starttoc{toc}
\vspace{4mm}
\hrule height 0.9pt
\makeatother
\endgroup

\newpage

\appendix
\addtocontents{toc}{\protect\setcounter{tocdepth}{1}}

\input{sec/X_suppl}

\end{document}

%% file: sec/0_abstract.tex
\begin{abstract}
Visual generation is increasingly ubiquitous in diverse domains, from text-to-image/video synthesis to multimodal interactive creation.
Yet prevailing monolithic models remain fundamentally constrained by their inability to learn cumulatively and evolve autonomously, which is a limitation we term the ``perpetual novice'' problem.
They lack mechanisms for structuring experience into reusable knowledge and therefore rely on brittle, ``from-scratch'' reasoning for each task, resulting in poor compositional generalization and inefficient knowledge retention.
Motivated by these limitations, we propose SymbOmni, an agentic omni-model designed for cumulative evolution through \emph{Symbolic Concept Learning}. At its core is the Symbolic Concept Box, an optimizable memory module that abstracts low-level operations into reusable Symbolic Workflow Instructions.
SymbOmni operates through an \emph{induction-transduction cycle}: experiences are abstracted into symbolic concepts (induction), which are then adaptively composed to solve novel tasks (transduction). The training is done by verbalized backpropagation with language-based feedback to enable continuous self-improvement without gradient-based model fine-tuning. Comprehensive experiments validate that (I) SymbOmni significantly outperforms existing agent-based systems for iterative creation and also surpasses closed-source models (\eg, Nano Banana, GPT-Image-1) in both image quality and task success rates; (II) SymbOmni effectively reduces token consumption by over 40\% while maintaining competitive generation quality; and (III) SymbOmni enables effective continual learning by achieving cumulative gains across multiple online-learning benchmarks and setting a new state of the art.
\end{abstract}

%% file: sec/1_intro.tex
\section{Introduction}
\label{sec:intro}

Visual generation has become essential to a wide range of applications, including text-to-image synthesis~\cite{rombach2022high,ho2020denoising,zhang2023adding}, text-to-video generation~\cite{hacohen2024ltx,kong2024hunyuanvideo}, and interactive content generation~\cite{brooks2023instructpix2pix,fu2024guiding}.
While standalone text-to-image or text-to-video models often struggle to meet complex creative demands~\cite{zhang2023adding,ye2023ip}, emerging unified or task-specific frameworks~\cite{ma2025janusflow,team2024chameleon,wang2024emu3} aim to bridge this gap. However, within the open-source community, these unified alternatives still suffer from unstable generation quality and insufficient structured planning~\cite{xue2025comfybench,huang2025comfygpt}. Although proprietary counterparts exhibit remarkably robust unified capabilities~\cite{gpt_image_1,de2024system,openai2025o3o4}, their closed nature inherently restricts architectural flexibility, downstream scalability, and broader multimodal adaptation~\cite{li2025perception,wu2025vila}.

Another fundamental limitation further challenges current visual generation paradigms. Existing large language models (LLMs) and agentic systems lack the capacity for cumulative learning (or continual learning) and self-evolution~\cite{guo2024large,chen2024agentverse,wu2023autogen}. By treating tasks in isolation, prevailing end-to-end architectures fail to distill useful and transferable knowledge from prior experiences~\cite{autogpt,wu2023autogen}. This results in the \emph{perpetual novice problem}, where the generative system exhibits poor compositional generalization, suboptimal knowledge retention, inefficient reasoning and brittle decision-making~\cite{wang2025mixture,liu2025breaking,xu2023towards}.

This perpetual novice problem is rooted at the framework level. Current prevailing unified generative-understanding paradigms often lack the mechanisms required to structure experiences into compact and reusable knowledge components. 
As a result, these systems must rely on from-scratch reasoning within a fixed parametric memory for each new task \cite{openai2025deep_research, openai2025o3o4}.
While alternative approaches may leverage external tools, they often produce unstable and non-reusable reasoning chains \cite{wu2023autogen, xie2023openagents, yao2022react}. Furthermore, existing attempts to mitigate these issues via implicit learning or tool-calling agents suffer from poor scalability and decision instability \cite{shinn2023reflexion, li2025torl}. Even workflow-based frameworks are hampered by rigid architectures that fail to adapt to dynamic user demands \cite{hong2024metagpt, qian2024chatdev, fan2025workflowllm}. Moreover, this limitation is particularly significant in enterprise scenarios with highly personalized workflows, where long-context methods can incur prohibitive computational costs and elevated hallucination risks \cite{xue2025comfybench, huang2025comfygpt, lewis2020retrieval}.

In this work, we introduce \textbf{\symbomni}, an agentic architecture designed for continual evolution through Symbolic Concept Learning. 
We argue that long-term adaptation requires agents to continuously abstract their experiences into structured, composable knowledge representations. To this end, SymbOmni incorporates a strong inductive bias that combines the flexibility of neural models with the rigor and interpretability of symbolic reasoning.
At the core of the architecture is the Symbolic Concept Box (CB), an optimizable memory that stores reusable units of knowledge, termed Symbolic Concepts. Each concept integrates semantic understanding with executable procedures that encode a successful problem-solving strategy. These strategies are further distilled into Symbolic Workflow Instructions (SWIs), which are dynamically retrieved, composed, and executed through a memory-augmented mechanism.
This design enables an \emph{Induction-Transduction cycle}. 
During induction, the agent abstracts past experiences into reusable symbolic concepts. During transduction, it retrieves, composes, and instantiates relevant concepts to solve new tasks. The cycle is completed through verbalized backpropagation~\cite{shinn2023reflexion,xiao2025verbalized,zhao2024marco,openai2025deep_research,yu2025generating,yuksekgonul2024textgrad}, which generates linguistic feedback to refine the agent’s symbolic knowledge and problem-solving strategies. In this way, SymbOmni supports continual self-improvement without requiring parameter fine-tuning. Our contributions are threefold:

\begin{itemize}[leftmargin=*, itemsep=0pt]
\setlength\itemsep{0.8em}
    \item \textbf{Continual adaptation framework}. SymbOmni is a Symbolic Concept Learning architecture for continual adaptation beyond monolithic models~\cite{wu2023autogen,xie2023openagents,hong2024metagpt}.
    \item \textbf{Verbalized learning}. SymbOmni integrates both inductive and transductive reasoning for cumulative learning via verbalized backpropagation~\cite{xiao2025verbalized,shinn2023reflexion}.
    \item  \textbf{Empirical effectiveness}. SymbOmni outperforms existing approaches, reducing token use by over 40\% and consistently improving workflows~\cite{xue2025comfybench,huang2025comfygpt,zheng2025deepresearcher}.
\end{itemize}

%% file: sec/2_formatting.tex
\begin{figure}[t!]
    \centering
    \begin{subfigure}[t]{0.535\linewidth}
        \centering
        \includegraphics[width=\linewidth]{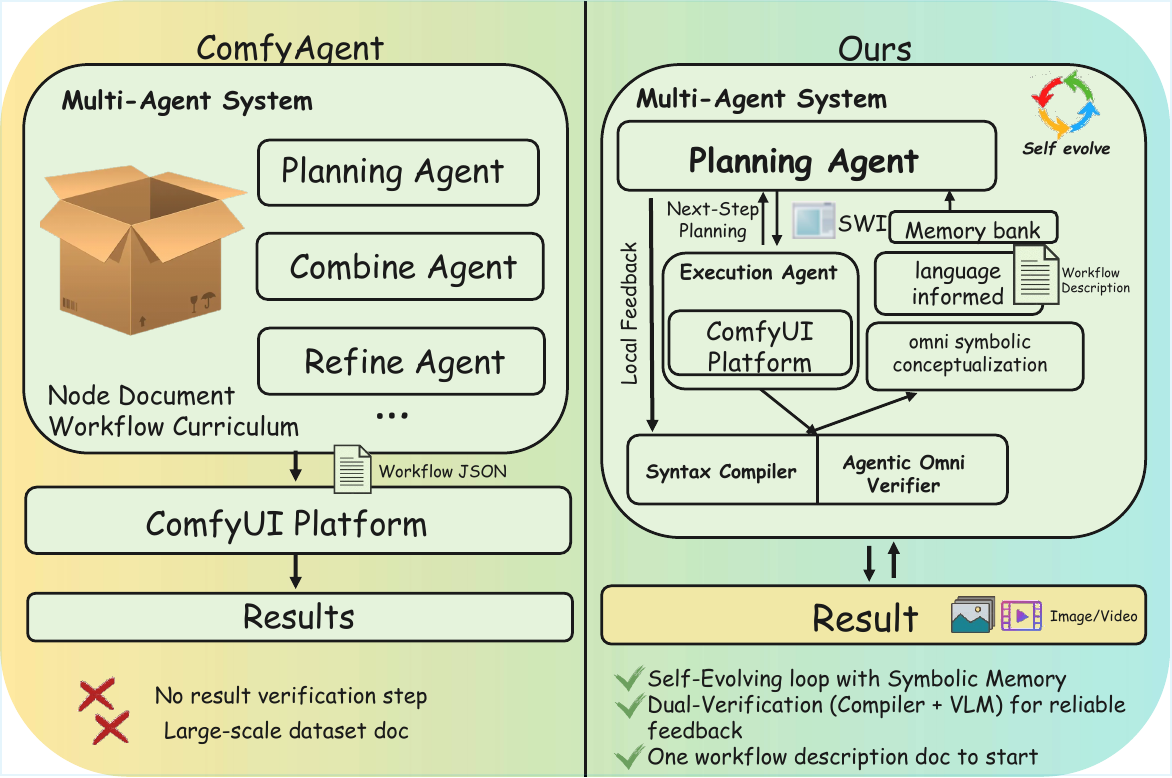}
        \caption{Comparison with baseline agent systems.}
        \label{fig:comparison_a}
    \end{subfigure}
    \hfill
    \begin{subfigure}[t]{0.45\linewidth}
        \centering
        \includegraphics[width=\linewidth]{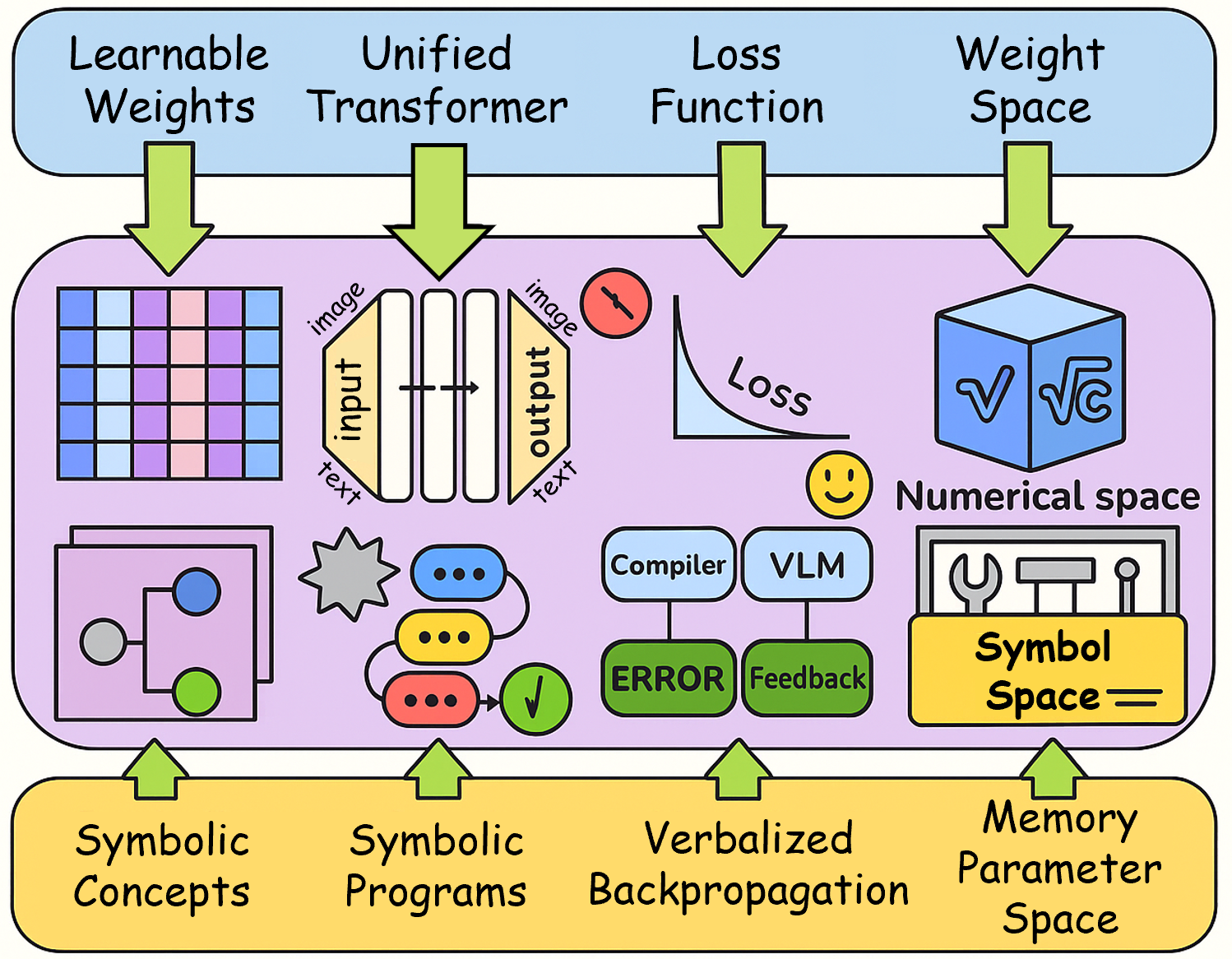}
        \caption{Unified models vs. our symbolic model.}
        \label{fig:comparison_b}
    \end{subfigure}
    \caption{(a) The comparison underscores the advantages of our model: a self-evolving capability through dual-verification feedback and an optimizable symbolic memory, which eliminates the dependency on large-scale pre-existing documentation required by the baseline approach. (b) Contrast between multimodal unified models and our agentic symbolic framework.}
    \label{fig:comparison}
\end{figure}

\section{Related Work}
\label{sec:relatedwork}

\paragraph{Agentic Systems and Workflow Automation.} Current AI systems often struggle with \textit{cumulative learning}, as they treat tasks in isolation and suffer from the perpetual novice problem, which leads to poor compositional generalization and inefficient knowledge retention~\cite{wu2023autogen,autogpt, wang2025mixture,liu2025breaking}. Recent work has sought to address these limitations through structured knowledge representations and automated workflow construction. Systems such as Deep Research~\cite{zheng2025deepresearcher} and OpenAI o3~\cite{openai2025o3o4} translate natural-language objectives into executable plans, while WorkflowLLM~\cite{fan2025workflowllm} and AFlow~\cite{zhang2025aflow} emphasize data-driven workflow optimization.
Early multi-agent systems~\cite{hong2024metagpt,qian2024chatdev} enable collaborative problem solving, but often rely on rigid interaction protocols. In specialized environments such as ComfyUI, approaches including ComfyAgent~\cite{xue2025comfybench} and ComfyGPT~\cite{huang2025comfygpt} support node-based workflow generation, and yet remain constrained by limited domain coverage and cascading error propagation. 
More general frameworks such as AutoGen~\cite{wu2023autogen} and ReAct~\cite{yao2022react} enable iterative refinement, but still lack reliable mechanisms for error diagnosis and recovery.
Building on these advances, our work formalizes an \textit{Induction-Transduction} cycle together with a dual-verifier architecture to enable genuine cumulative learning.

\paragraph{Visual Generation and Reasoning Systems.}
Visual generation has progressed from task-specific models~\cite{zhang2023adding,ye2023ip} to general-purpose multimodal systems~\cite{ma2025janusflow,team2024chameleon}, driven by advances in diffusion models~\cite{rombach2022high} and unified architectures such as GPT-Image-1~\cite{gpt_image_1}.
In parallel, large reasoning models have substantially improved planning through System-2 reasoning~\cite{jaech2024openai} and reinforcement learning~\cite{guo2025deepseek}. However, these systems often depend on lengthy and computationally expensive contexts, which can amplify hallucinations and limit efficient knowledge reuse.
As illustrated in Figure~\ref{fig:comparison_b}, existing unified multimodal models primarily acquire knowledge through static parametric updates in weight space. In contrast, our architecture introduces an explicit symbolic concept layer that continuously abstracts, consolidates, and reuses knowledge through verbalized learning. SymbOmni unifies symbolic concept learning with visual workflow generation, yielding reusable knowledge components that reduce computational overhead, support continual improvement, and address the challenge of structured knowledge consolidation beyond conventional end-to-end approaches.

%% file: sec/3_finalcopy.tex
\section{\symbomni: Evolving Agentic Visual Generation}
\label{sec:methodology}

\subsection{Preliminaries on ComfyUI and Node-based Workflows}
\label{sec:preliminary}

ComfyUI~\cite{comfyui} is an open-source, node-based interface for visual generation. It can represent generative pipelines as directed acyclic graphs (DAGs) of interconnected nodes, each encapsulating a discrete operation (\eg, model loading, latent sampling, or post-processing). Users compose multi-stage workflows by wiring node outputs to inputs, and the result is serialized as a JSON file specifying all node types, parameters, and connections. This modularity offers fine-grained control of generated content but demands significant expertise in node selection, configuration, and connectivity. To address this challenge, our work focuses on automating the construction of such workflows.

\subsection{Overview of the SymbOmni Framework}
\label{sec:theoretical_foundation}

Our Omni Agentic Model introduces a unified framework that integrates induction and transduction within a self-reinforcing cycle. During the transductive phase, the system draws on reusable symbolic knowledge stored in the Symbolic Concept Box to solve novel tasks. It adapts abstract concepts and constraints to specific problem instances while preserving the underlying knowledge base. The inductive phase serves as the primary learning mechanism, transforming concrete experiences (both successful and unsuccessful queries) into symbolic knowledge by extracting generalizable rules and patterns from individual cases. This process continuously expands and refines the Symbolic Concept Box.
Put together, these mechanisms effectively form the coherent ``Induction-Memory-Transduction-Experience'' cycle illustrated in Figure~\ref{fig:framework}. Induction enriches symbolic memory with generalized knowledge, which subsequently improves the effectiveness of transduction on future tasks. This reciprocal reinforcement supports continuous meta-learning, allowing each new task and experience to further strengthen the system's reasoning capabilities.

\subsection{Symbolic Concepts and Concept Box}
\label{sec:core_components}

A key characteristic of our framework is its representation of knowledge in the form of Symbolic Concepts. Each Symbolic Concept, denoted by $C_k$, encapsulates a reusable unit of expert knowledge and is defined by the quadruple:
\begin{equation}
C_k = (Desc_k, SWI_k, Params_k, Score_k)
\label{eq:symbolic_concept}
\end{equation}
where $ Desc_k $ is a natural-language description that specifies the concept's intended purpose; $ SWI_k $ is a Symbolic Workflow Instruction, represented as a parameterized template describing a sequence of operations; $ Params_k $ contains optimized parameter configurations learned from prior applications; and $ Score_k $ records a dynamically updated measure of the concept's historical effectiveness. During execution, the parameter slots in $ SWI_k $ are instantiated using the corresponding values stored in $ Params_k $. The collection of all Symbolic Concepts forms the Concept Box, which serves as the system's long-term, optimizable memory. 
We formalize the CB as a language system, $ \mathcal{L} = (\Sigma, \mathcal{R}) $, where $ \Sigma = \{Desc_k\} $ defines a vocabulary of semantic primitives, and $ \mathcal{R} = \{SWI_k\} $ denotes a set of production rules that specify valid workflow compositions. Under this formulation, planning can be viewed as a grammatical derivation in $ \mathcal{L} $, transforming problem solving from the derivation of solutions from scratch into the composition of previously validated building blocks. The natural-language descriptions $Desc_k$ provide the semantic foundation for assessing the similarity between stored concepts and new tasks, thereby supporting the hierarchical retrieval process detailed in Section~\ref{sec:retrieval_planning}.

\subsection{The Self-Evolution Cycle}
\label{sec:evolution_loop}

\begin{figure*}[t]
    \centering
    \includegraphics[width=1\linewidth]{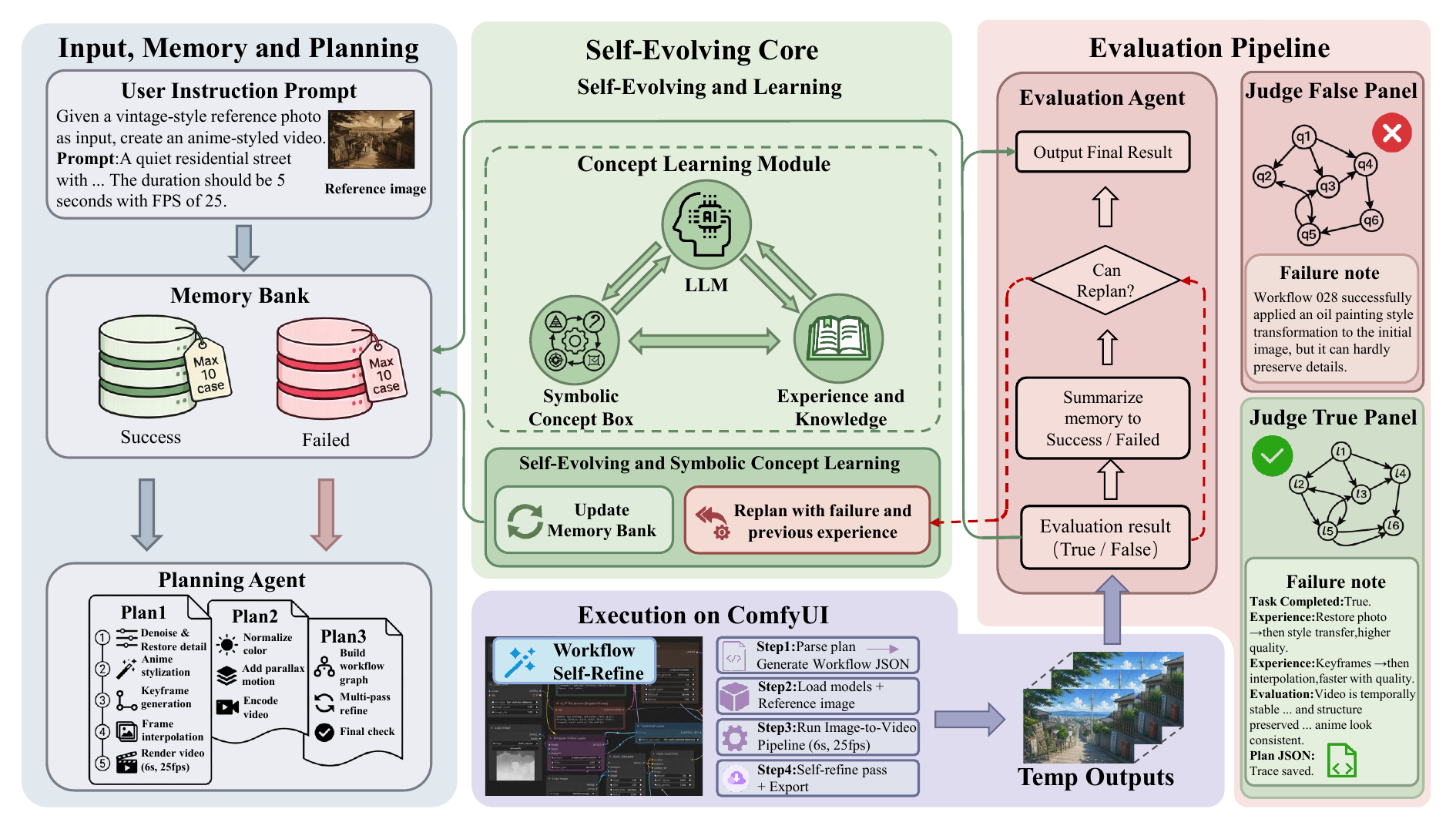}
    \caption{Illustration of SymbOmni's self-evolving symbolic concept learning mechanism. The framework iteratively transforms user instructions into solutions through a dual-feedback loop: successful outcomes are abstracted into symbolic experiences and stored in the Memory Bank, while failures trigger a replay process that refines future planning. This continuous learning cycle enables the agent to accumulate and leverage transferable knowledge beyond individual tasks.}
    \label{fig:framework}
\end{figure*}

Equipped with the structured Symbolic CB, SymbOmni operates through a continuous four-phase cycle that seamlessly integrates planning, execution, and learning, as illustrated in Figure~\ref{fig:framework}. In this cycle, we introduce a dual-feedback mechanism that enables the system to accumulate, refine, and reuse knowledge from previous experience. The following sections describe each phase in detail.

\paragraph{\underline{Phase 1: Transduction (Planning).}}
\label{sec:retrieval_planning}
The cycle begins with the Transduction phase (\ie, task planning), where the system translates a new task instruction into a concrete, executable plan by leveraging knowledge stored in the CB. Given task instruction $I_{\text{new}}$, the hierarchical retrieval proceeds as follows. First, $ST = \text{Decompose}(I_{\text{new}})$ parses the instruction into semantic subtasks with dependencies. Then, concepts are retrieved through a two-level cascade:
\begin{equation}
    C_{\text{ret}} = \text{Rank}\circ\text{Filter}_{\text{struct}}\circ\text{Retrieve}_{\text{sem}}(ST, \text{CB})
\end{equation}
\noindent
where the $\text{Retrieve}_{\text{sem}}$ operation identifies relevant concepts based on the cosine similarity between their embeddings, \ie, $\cos(\text{Embed}(Desc_k), \text{Embed}(ST_m))$. %
The $\text{Filter}_{\text{struct}}$ operation then enforces dependency constraints, while $\text{Rank}$ orders the remaining concepts according to their utility scores.
Conditioned on the retrieved concepts and the grammar $\mathcal{G}$, the LLM planner generates workflows:
\begin{equation}
WF_{\text{cand}} = \text{Instantiate}(\text{LLM}(I_{\text{new}} | C_{\text{ret}}, \mathcal{G}))
\end{equation}
where $\text{Instantiate}$ binds the parameters $\text{Params}_k$ to the corresponding workflow templates $SWI_k$, producing executable workflow candidates.

\paragraph{\underline{Phase 2: Execution.}}
The Execution phase is quite straightforward. The system executes the planned workflow $WF_{\text{cand}}$ step-by-step, invoking the corresponding tools (\eg, image generators, filters) to produce the final output $O$. 
Throughout this process, the complete execution trajectory $\tau$ is recorded in detail and subsequently used as supervision for the learning phase:
\begin{equation}
    \tau = (I_{\text{new}}, WF_{\text{cand}}, O, C_{\text{ret}}, \text{ExecutionLog}).
\end{equation}

\paragraph{\underline{Phase 3: Induction (Learning) and Evaluation.}}
\label{sec:induction_learning}
The Induction phase forms the core of the system's learning process. 
It begins with a critical evaluation step, in which the Evaluation Agent assesses the generated output $O$ against the input instruction $I_{\text{new}}$ and produces a binary judgment, $\text{Judge} \in \{\text{True}, \text{False}\}$. This judgment determines the subsequent learning path, thereby leading to the proposed dual-feedback mechanism.

If $\text{Judge} = \text{True}$ (successful execution), indicating successful execution, the trajectory $\tau$ is distilled into positive symbolic experience. Through \emph{symbolic abstraction}, the system extracts generalizable knowledge from the trajectory, either constructing new composite concepts or reinforcing existing ones.
\begin{equation}
    \text{CB} \leftarrow \text{CB} \cup \text{Symbolize}(\tau, \text{Positive})
\end{equation}

If $\text{Judge} = \text{False}$ (failed execution), indicating failed execution, the system initiates a detailed diagnostic process. Inspired by \cite{xiao2025verbalized}, the failure is analyzed through \emph{verbalized backpropagation}. 
Specifically, two complementary verifiers evaluate the execution: $\text{Error}_{\text{hard}}$ assesses syntactic and procedural correctness, while $\text{Feedback}_{\text{soft}}$ evaluates semantic quality. %
An LLM then analyzes these signals and feeds them into a structured \emph{linguistic loss} $L_{\text{lang}}$, providing an interpretable diagnosis of the failure. 
Subsequently, chain attribution computes \textit{symbolic gradients} $\nabla_{\text{sym}}$ to pinpoint refinement targets in the CB, guiding a replay and refinement process.
\begin{equation}
    L_{\text{lang}} = \text{LLM}(\mathcal{P}_{\text{loss}}(\tau, \text{Error}_{\text{hard}}, \text{Feedback}_{\text{soft}}))
\end{equation}
\begin{equation}
    \text{CB} \leftarrow \text{Refine}(\text{CB}, \text{Replay}(\tau, \nabla_{\text{sym}}))
\end{equation}

\paragraph{\underline{Phase 4: Memory Optimization.}}
In the \emph{Memory Optimization} phase, the insights acquired during Phase 3 are consolidated into the Symbolic CB. Newly constructed or refined symbolic concepts are incorporated into memory. Successful trajectories are abstracted into reusable knowledge, denoted by $C_{\text{success}}$, whereas failed trajectories yield either negative concepts for avoiding recurring errors, $C_{\text{negative}}$, or targeted parameter updates, $C_{\text{refine}}$. Guided by verbalized learning, this update completes a full self-evolution cycle. The resulting enriched CB enables the system to solve related tasks more effectively in the future, thereby effectively achieving continual learning.

\section{Experiments and Results}

We evaluate our system's performance on three complementary public benchmarks: ComfyBench for assessing AI agent workflows and automatic workflow generation from natural language descriptions~\cite{xue2025comfybench}, GenEval for assessing semantic consistency and visual fidelity in text-to-image generation tasks~\cite{ghosh2023geneval}, and ReasonEdit for examining performance in multi-step reasoning and complex image editing scenarios~\cite{huang2024smartedit}. Additionally, we analyze the token efficiency and runtime of SymbOmni on large-scale tasks, and examine the performance differences between Online and Offline modes to demonstrate self-evolution capabilities and token efficiency~\cite{huang2025comfygpt, shinn2023reflexion, zheng2025deepresearcher, liu2025advances, wang2025mixture}. We further conduct ablation studies to validate the contribution of the symbolic concept memory, a large-scale user preference study to evaluate perceptual quality, and out-of-domain generalization experiments to assess transferability of the learned concepts.

\subsection{Experimental Setup}

To rigorously evaluate planning and learning capabilities, we enhance the workflow library with improved coordination mechanisms and high-quality workflows, while deliberately retaining several suboptimal ones. All experiments were conducted in the ComfyUI framework~\cite{comfyui}, using Gemini 2.5 Flash~\cite{comanici2025gemini} as the reasoning engine. For fair comparison across agentic systems, we standardize key hyperparameters, setting the maximum search depth to 10 and the maximum number of retries to 4~\cite{yao2022react,patil2024gorilla,li2025search,schick2023toolformer}. Unified generative models and published baselines followed their respective benchmark protocols. We evaluate performance along four different dimensions: \emph{Pass Rate}, the proportion of generated workflows that execute successfully with the correct structure; \emph{Resolve Rate}, the proportion of final outputs that satisfy the semantic requirements of the instruction; \emph{Generation Quality}, the semantic consistency and perceptual quality of the outputs; and \emph{Token Consumption}, the average number of tokens used per task as an indicator of reasoning efficiency~\cite{liu2024agentbench,qiao2025benchmarking,niu2025wise}.

\subsection{Qualitative Comparison}

\begin{figure*}
    \centering
    \includegraphics[width=1\linewidth]{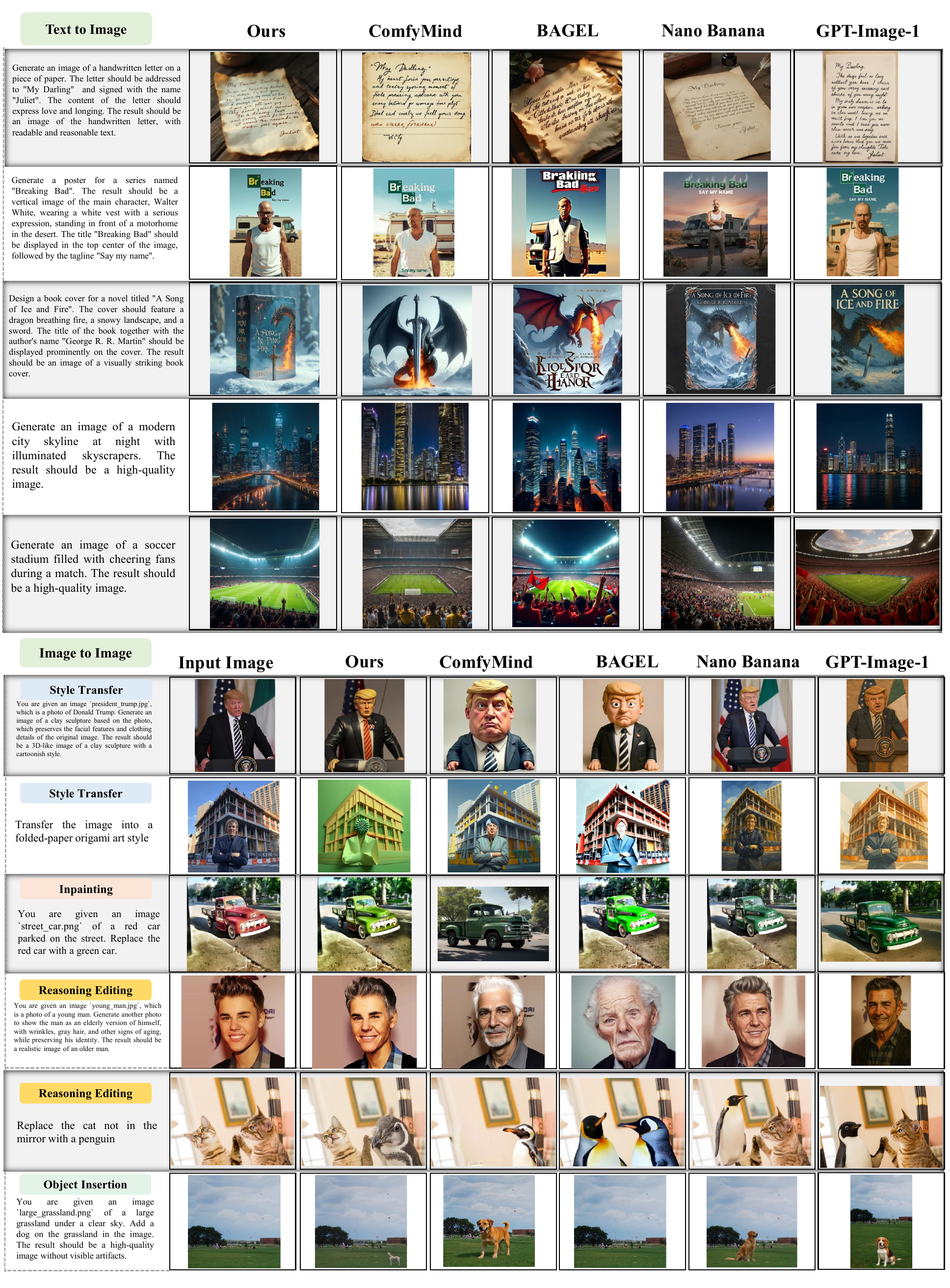}
    \caption{Qualitative comparison among SymbOmni, ComfyMind, Bagel, Nano Banana and GPT-Image-1 for two representative visual generation tasks. Top five rows: text-to-image generation; Bottom five rows: image-to-image reasoning and generation. We can observe that the quality of SymbOmni's output is on par with the state-of-the-art proprietary image generation systems, and SymbOmni shows the best instruction-following ability among all.}
    \label{fig:qualitative_results}
\end{figure*}

We qualitatively compare SymbOmni with state-of-the-art collaborative AI systems~\cite{guo2026comfymind} and leading closed-source unified models~\cite{gpt_image_1,deng2025emerging,comanici2025gemini}.
As shown in Figure~\ref{fig:qualitative_results}, our method demonstrates clear advantages in handling complex compositional tasks.
In the text-to-image generation tasks that require precise adherence to complex instructions, SymbOmni achieves stronger semantic alignment when instructions contain multiple fine-grained requirements. For the ``Breaking Bad'' poster, it correctly depicts Walter White wearing a white vest with a serious expression in a desert setting, while accurately rendering the title and tagline. In contrast, several other baselines fail to preserve the character appearance, scene details, or textual elements.
In the complex image editing and image-to-image generation tasks, 
SymbOmni reliably executes multi-step editing operations. In the ``red car to green car'' example, it both changes the car color and extends the canvas as instructed, whereas other methods often satisfy only part of the request. Similarly, in transforming Donald Trump into a clay sculpture, SymbOmni better preserves facial identity while applying the intended cartoon-like, three-dimensional clay texture. In contrast, baselines frequently distort the subject or fail to reproduce the desired style.
These results suggest that symbolic concept learning enables SymbOmni to interpret and execute complex, multi-faceted instructions more accurately, yielding stronger performance across both image generation and editing tasks compared to existing methods~\cite{zhang2023adding,qiu2023controlling,ye2023ip,brooks2023instructpix2pix,fu2024guiding}.

\subsection{Quantitative Experiments}
\label{sec:qualitative_experiment}

\begin{table*}[t]
\centering
\setlength{\abovecaptionskip}{6pt}
  \setlength{\belowcaptionskip}{2pt}
\renewcommand{\arraystretch}{1.2}
\renewcommand{\tabcolsep}{3.5pt}
\resizebox{\linewidth}{!}{
\begin{tabular}{l|cccccccc}
\Xhline{1.2pt}
\multirow{2}{*}{\textbf{Agent}} & \multicolumn{2}{c}{\textbf{Vanilla}} & \multicolumn{2}{c}{\textbf{Complex}} & \multicolumn{2}{c}{\textbf{Creative}} & \multicolumn{2}{c}{\textbf{Total}} \\
\cmidrule(lr){2-3} \cmidrule(lr){4-5} \cmidrule(lr){6-7} \cmidrule(lr){8-9}
& \%Pass{\red{$\uparrow$}}  & \%Resolve{\red{$\uparrow$}} & \%Pass{\red{$\uparrow$}} & \%Resolve{\red{$\uparrow$}} & \%Pass{\red{$\uparrow$}} & \%Resolve{\red{$\uparrow$}} & \%Pass{\red{$\uparrow$}} & \%Resolve{\red{$\uparrow$}} \\
\Xhline{1.2pt}
GPT-4o + Few-shot~\cite{brown2020language} & 32.0 & 27.0 & 16.7 & 8.3 & 7.5 & 0.0 & 22.5 & 16.0 \\
GPT-4o + CoT~\cite{wei2022chain} & 44.0{\tiny\red{$\uparrow$12.0}} & 29.0{\tiny\red{$\uparrow$2.0}} & 11.7{\tiny\red{$\downarrow$5.0}} & 8.3{\tiny\red{0.0}} & 12.5{\tiny\red{$\uparrow$5.0}} & 0.0{\tiny\red{0.0}} & 28.0{\tiny\red{$\uparrow$5.5}} & 17.0{\tiny\red{$\uparrow$1.0}} \\
GPT-4o + CoT-SC~\cite{wang2022self} & 45.0{\tiny\red{$\uparrow$13.0}} & 34.0{\tiny\red{$\uparrow$7.0}} & 11.7{\tiny\red{$\downarrow$5.0}} & 5.0{\tiny\red{$\downarrow$3.3}} & 15.0{\tiny\red{$\uparrow$7.5}} & 0.0{\tiny\red{0.0}} & 29.0{\tiny\red{$\uparrow$6.5}} & 18.5{\tiny\red{$\uparrow$2.5}} \\
Claude-3.5-Sonnet + RAG~\cite{lewis2020retrieval} & 27.0{\tiny\red{$\downarrow$5.0}} & 13.0{\tiny\red{$\downarrow$14.0}} & 23.0{\tiny\red{$\uparrow$6.3}} & 6.7{\tiny\red{$\downarrow$1.6}} & 7.5{\tiny\red{0.0}} & 0.0{\tiny\red{0.0}} & 22.0{\tiny\red{$\downarrow$0.5}} & 8.5{\tiny\red{$\downarrow$7.5}} \\
Llama-3.1-70B + RAG & 58.0{\tiny\red{$\uparrow$26.0}} & 32.0{\tiny\red{$\uparrow$5.0}} & 23.0{\tiny\red{$\uparrow$6.3}} & 10.0{\tiny\red{$\uparrow$1.7}} & 15.0{\tiny\red{$\uparrow$7.5}} & 5.0{\tiny\red{$\uparrow$5.0}} & 39.0{\tiny\red{$\uparrow$16.5}} & 20.0{\tiny\red{$\uparrow$4.0}} \\
GPT-4o + RAG & 62.0{\tiny\red{$\uparrow$30.0}} & 41.0{\tiny\red{$\uparrow$14.0}} & 45.0{\tiny\red{$\uparrow$28.3}} & 21.7{\tiny\red{$\uparrow$13.4}} & \blueboxtext{40.0}{\tiny\red{$\uparrow$32.5}} & 7.5{\tiny\red{$\uparrow$7.5}} & 52.0{\tiny\red{$\uparrow$29.5}} & 23.0{\tiny\red{$\uparrow$7.0}} \\
o1-mini + RAG & 32.0{\tiny\red{0.0}} & 16.0{\tiny\red{$\downarrow$11.0}} & 21.7{\tiny\red{$\uparrow$5.0}} & 8.3{\tiny\red{0.0}} & 12.5{\tiny\red{$\uparrow$5.0}} & 7.5{\tiny\red{$\uparrow$7.5}} & 25.0{\tiny\red{$\uparrow$2.5}} & 12.0{\tiny\red{$\downarrow$4.0}} \\
o1-preview + RAG & \blueboxtext{70.0}{\tiny\red{$\uparrow$38.0}} & \blueboxtext{46.0}{\tiny\red{$\uparrow$19.0}} & \blueboxtext{48.3}{\tiny\red{$\uparrow$31.6}} & \blueboxtext{23.3}{\tiny\red{$\uparrow$15.0}} & 30.0{\tiny\red{$\uparrow$22.5}} & \blueboxtext{12.5}{\tiny\red{$\uparrow$12.5}} & \blueboxtext{55.5}{\tiny\red{$\uparrow$33.0}} & \blueboxtext{32.5}{\tiny\red{$\uparrow$16.5}} \\
\midrule
ComfyAgent~\cite{xue2025comfybench} & 67.0 & 46.0 & 48.3 & 21.7 & 40.0 & 15.0 & 56.0 & 32.5 \\
ComfyMind~\cite{guo2026comfymind} (reproduced) & 100.0{\tiny\red{$\uparrow$33.0}} & 84.0{\tiny\red{$\uparrow$38.0}} & 100.0{\tiny\red{$\uparrow$51.7}} & 75.0{\tiny\red{$\uparrow$53.3}} & 100.0{\tiny\red{$\uparrow$60.0}} & 42.5{\tiny\red{$\uparrow$27.5}} & 100.0{\tiny\red{$\uparrow$44.0}} & 73.0{\tiny\red{$\uparrow$40.5}} \\
\textbf{SymbOmni (Ours)} & \blueboxtext{100.0}{\tiny\red{$\uparrow$33.0}} & \blueboxtext{95.0}{\tiny\red{$\uparrow$49.0}} & \blueboxtext{100.0}{\tiny\red{$\uparrow$51.7}} & \blueboxtext{83.3}{\tiny\red{$\uparrow$61.6}} & \blueboxtext{100.0}{\tiny\red{$\uparrow$60.0}} & \blueboxtext{67.5}{\tiny\red{$\uparrow$52.5}} & \blueboxtext{100.0}{\tiny\red{$\uparrow$44.0}} & \blueboxtext{86.0}{\tiny\red{$\uparrow$53.5}} \\
\midrule
\multicolumn{9}{l}{\textit{+ Nano Banana~\cite{comanici2025gemini} (Unified SOTA Model)}} \\
ComfyMind + Nano Banana & 100.0{\tiny\red{$\uparrow$33.0}} & 97.0{\tiny\red{$\uparrow$51.0}} & 100.0{\tiny\red{$\uparrow$51.7}} & 80.0{\tiny\red{$\uparrow$58.3}} & 100.0{\tiny\red{$\uparrow$60.0}} & 57.5{\tiny\red{$\uparrow$42.5}} & 100.0{\tiny\red{$\uparrow$44.0}} & 84.0{\tiny\red{$\uparrow$51.5}} \\
\textbf{SymbOmni + Nano Banana} & \blueboxtext{100.0}{\tiny\red{$\uparrow$33.0}} & \blueboxtext{99.0}{\tiny\red{$\uparrow$53.0}} & \blueboxtext{100.0}{\tiny\red{$\uparrow$51.7}} & \blueboxtext{88.3}{\tiny\red{$\uparrow$66.6}} & \blueboxtext{100.0}{\tiny\red{$\uparrow$60.0}} & \blueboxtext{75.0}{\tiny\red{$\uparrow$60.0}} & \blueboxtext{100.0}{\tiny\red{$\uparrow$44.0}} & \blueboxtext{91.0}{\tiny\red{$\uparrow$58.5}} \\
\Xhline{1.2pt}
\end{tabular}
}
\caption{Evaluation of autonomous workflow construction on ComfyBench~\cite{xue2025comfybench}. We highlight the best results with colored boxes.}
\label{tab:comfybench}
\end{table*}

ComfyBench evaluates the ability to construct executable workflows from task descriptions~\cite{xue2025comfybench}. As shown in Table~\ref{tab:comfybench}, SymbOmni achieves a perfect 100\% pass rate and a total resolve rate of 86.0\%, significantly outperforming all baseline methods. The advantage is most substantial on complex and creative tasks. For complex tasks, the resolve rate improves from 75.0\% to 83.3\% (11.1\% relative gain). For the challenging creative tasks, it improves from 42.5\% to 67.5\%, a remarkable 58.8\% relative gain. These gains stem from SymbOmni's ability to accumulate and reuse learned symbolic concepts, enabling more flexible workflow composition and stronger generalization to novel task configurations~\cite{niu2025flow, zhang2025aflow, fan2025workflowllm}.

When further equipped with a unified SOTA model Nano Banana~\cite{comanici2025gemini}, SymbOmni reaches 91.0\% total resolve rate. Under identical tool configurations, SymbOmni consistently surpasses ComfyMind (91.0\% vs.\ 84.0\% overall; 75.0\% vs.\ 57.5\% on Creative), verifying that the gains originate from our symbolic concept learning mechanism rather than the underlying generative model.

\begin{figure}[t]
\centering

\begin{minipage}[b]{0.35\textwidth}
\centering
\includegraphics[width=\linewidth]{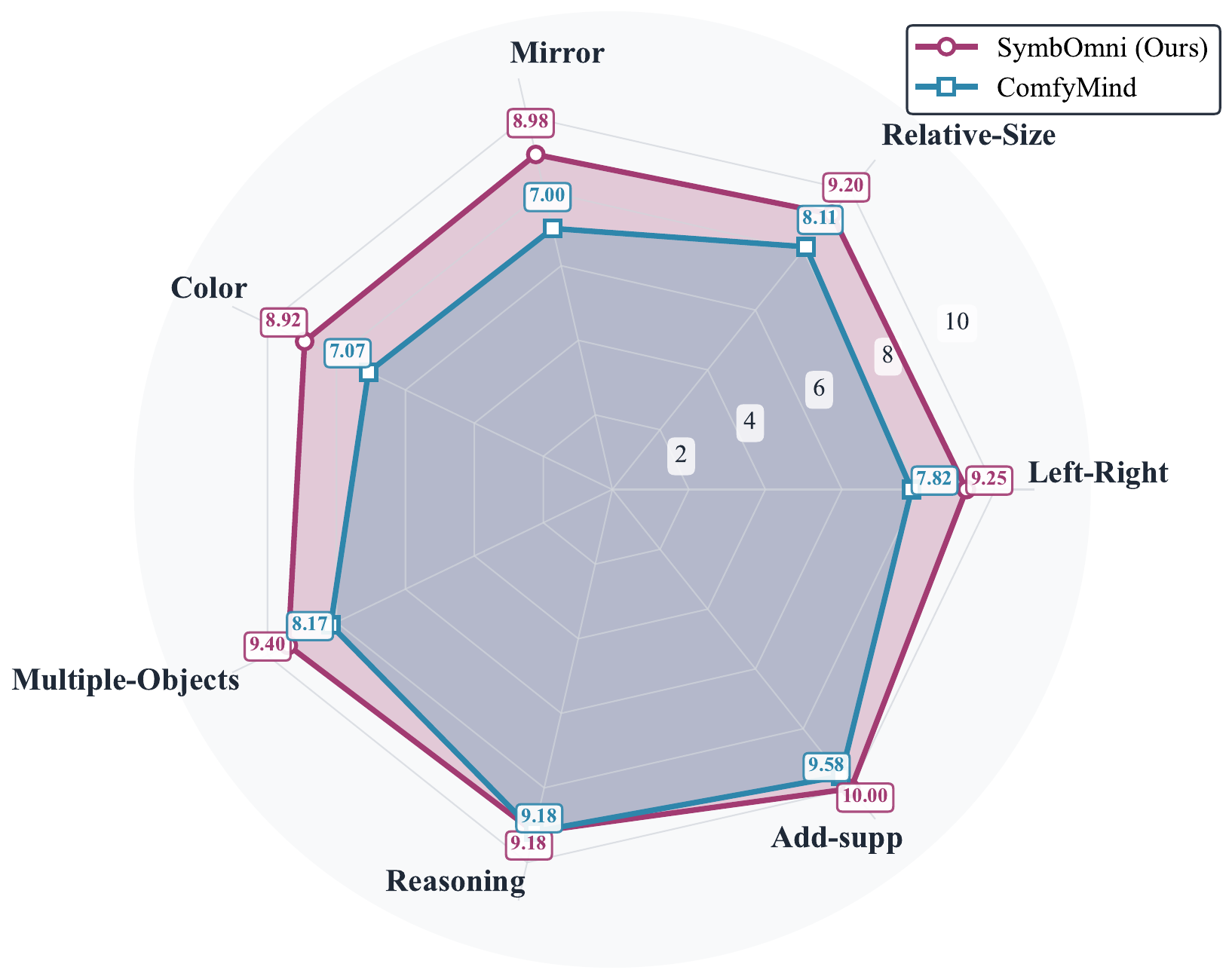}
\subcaption{Comparison of ComfyMind and SymbOmni on ReasonEdit.}
\label{fig:reasonedit_radar}
\end{minipage}
\hfill
\begin{minipage}[b]{0.60\textwidth}
\centering
\definecolor{comfymind}{RGB}{180, 120, 50}
\definecolor{nanob}{RGB}{100, 160, 200}
\definecolor{symbomni}{RGB}{50, 115, 180}

\resizebox{\linewidth}{!}{
\begin{tikzpicture}
  \begin{axis}[
      ybar=2pt, 
      bar width=7pt, 
      width=14cm,
      height=6cm,
      ymin=6.5, ymax=10.5, 
      ylabel={Score (0-10)},
      ylabel style={font=\small},
      xlabel={Task Category},
      xlabel style={font=\small},
      xtick=data,
      xticklabels={Left-Right, Relative-Size, Mirror, Color, Multiple-Obj, Reasoning, Add-supp},
      xticklabel style={font=\scriptsize, rotate=30, anchor=north east}, %
      ytick={7,8,9,10},
      legend style={at={(0.5,1.05)}, anchor=south, legend columns=3, font=\scriptsize, draw=none},
      grid=major,
      grid style={dashed, gray!30},
      every node near coord/.append style={
          font=\scriptsize,
          rotate=90,     
          anchor=west,     
          /pgf/number format/fixed,
          /pgf/number format/precision=2
      }
  ]

  \addplot[
      comfymind, fill=comfymind,
      nodes near coords %
  ] coordinates {
      (1,7.821) (2,8.114) (3,7.000) (4,7.074) (5,8.167) (6,9.175) (7,9.583)
  };

  \addplot[
      nanob, fill=nanob,
      nodes near coords
  ] coordinates {
      (1,9.804) (2,9.818) (3,8.383) (4,9.778) (5,9.563) (6,9.158) (7,10.000)
  };

  \addplot[
      symbomni, fill=symbomni,
      nodes near coords
  ] coordinates {
      (1,9.250) (2,9.204) (3,8.983) (4,8.923) (5,9.396) (6,9.183) (7,10.000)
  };

  \legend{ComfyMind, Nano Banana, SymbOmni}
  \end{axis}
\end{tikzpicture}
}
\subcaption{Detailed subcategory comparison with Nano Banana.}
\label{fig:reasonedit_comparison}
\end{minipage}

\caption{Evaluation of reasoning image-to-image editing on ReasonEdit~\cite{huang2024smartedit}. (a) Radar chart showing overall performance comparison of SymbOmni and ComfyMind across subcategories. (b) Detailed bar chart comparing SymbOmni, Nano Banana~\cite{comanici2025gemini}, and ComfyMind~\cite{guo2026comfymind} across seven reasoning-intensive task types.}
\label{fig:reasonedit}
\end{figure}
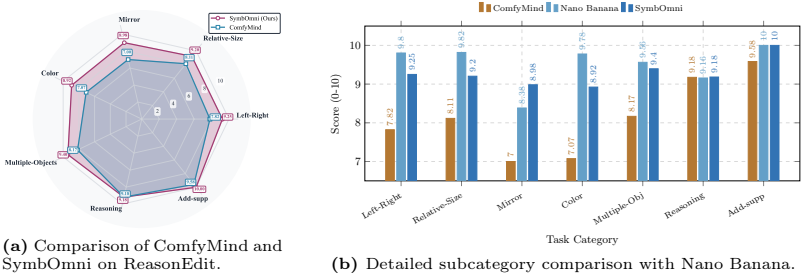

\paragraph{Reason-Edit: Complex Visual Reasoning Editing.}
The ReasonEdit benchmark evaluates multi-step visual reasoning editing capabilities~\cite{huang2024smartedit}. As shown in Figure~\ref{fig:reasonedit_radar}, SymbOmni achieves an overall score of 9.190, substantially outperforming ComfyMind (8.135), with the most pronounced gains in complex spatial reasoning (+1.980 on Mirror, +1.229 on Multiple-Objects)~\cite{liu2025breaking, xu2023towards, wang2024rethinking}.

Figure~\ref{fig:reasonedit_comparison} compares SymbOmni with unified multimodal models. SymbOmni performs particularly well on reasoning-intensive tasks, achieving a score of 8.983 on the Mirror task, which is substantially higher than Nano Banana (8.383; +7.2\%) and ComfyMind (7.000; +28.3\%).
This validates the effectiveness of symbolic workflow decomposition for multi-step spatial transformations. Although Nano Banana performs slightly better in some subcategories (\eg, Left-Right and Color), these gains largely reflect the capabilities of its underlying generative model. As an agentic system, SymbOmni is inherently constrained by the quality of its constituent tools; nevertheless, it achieves a perfect score of 10.000 on Add-supp and remains competitive across all evaluation dimensions.

\begin{table}[!t]
\centering
\renewcommand{\arraystretch}{1.2}
\renewcommand\tabcolsep{3.5pt}
\setlength{\abovecaptionskip}{6pt}
\setlength{\belowcaptionskip}{5pt}
\resizebox{\linewidth}{!}{
\begin{tabular}{l|c|cccccc}
\Xhline{1.2pt}
\rowcolor{CadetBlue!20}
\textbf{Method} & \textbf{Overall}{\red{$\uparrow$}} & \textbf{Single Obj.}{\red{$\uparrow$}} & \textbf{Two Obj.}{\red{$\uparrow$}} & \textbf{Counting}{\red{$\uparrow$}} & \textbf{Colors}{\red{$\uparrow$}} & \textbf{Position}{\red{$\uparrow$}} & \textbf{Attr. Binding}{\red{$\uparrow$}} \\
\Xhline{1.2pt}
\multicolumn{8}{l}{\textit{Frozen Text Encoder Mapping Methods}} \\
SDv1.5~\cite{rombach2022high} & 0.43 & 0.97 & 0.38 & 0.35 & 0.76 & 0.04 & 0.06 \\
SDv2.1~\cite{rombach2022high} & 0.50{\tiny\red{$\uparrow$0.07}} & 0.98{\tiny\red{$\uparrow$0.01}} & 0.51{\tiny\red{$\uparrow$0.13}} & 0.44{\tiny\red{$\uparrow$0.09}} & 0.85{\tiny\red{$\uparrow$0.09}} & 0.07{\tiny\red{$\uparrow$0.03}} & 0.17{\tiny\red{$\uparrow$0.11}} \\
SD-XL~\cite{podell2024sdxl} & 0.55{\tiny\red{$\uparrow$0.12}} & 0.98{\tiny\red{$\uparrow$0.01}} & 0.74{\tiny\red{$\uparrow$0.36}} & 0.39{\tiny\red{$\uparrow$0.04}} & 0.85{\tiny\red{$\uparrow$0.09}} & 0.15{\tiny\red{$\uparrow$0.11}} & 0.23{\tiny\red{$\uparrow$0.17}} \\
DALLE-2~\cite{ramesh2022hierarchical} & 0.52{\tiny\red{$\uparrow$0.09}} & 0.94{\tiny\blue{$\downarrow$0.03}} & 0.66{\tiny\red{$\uparrow$0.28}} & 0.49{\tiny\red{$\uparrow$0.14}} & 0.77{\tiny\red{$\uparrow$0.01}} & 0.10{\tiny\red{$\uparrow$0.06}} & 0.19{\tiny\red{$\uparrow$0.13}} \\
SD3-Medium~\cite{esser2024scaling} & \blueboxtext{0.74}{\tiny\red{$\uparrow$0.31}} & \blueboxtext{0.99}{\tiny\red{$\uparrow$0.02}} & \blueboxtext{0.94}{\tiny\red{$\uparrow$0.56}} & \blueboxtext{0.72}{\tiny\red{$\uparrow$0.37}} & \blueboxtext{0.89}{\tiny\red{$\uparrow$0.13}} & \blueboxtext{0.33}{\tiny\red{$\uparrow$0.29}} & \blueboxtext{0.60}{\tiny\red{$\uparrow$0.54}} \\

\midrule
\multicolumn{8}{l}{\textit{Multimodal Unified Models}} \\
LlamaGen~\cite{sun2024autoregressive} & 0.32 & 0.71 & 0.34 & 0.21 & 0.58 & 0.07 & 0.04 \\
LWM~\cite{liu2025world} & 0.47{\tiny\red{$\uparrow$0.15}} & 0.93{\tiny\red{$\uparrow$0.22}} & 0.41{\tiny\red{$\uparrow$0.07}} & 0.46{\tiny\red{$\uparrow$0.25}} & 0.79{\tiny\red{$\uparrow$0.21}} & 0.09{\tiny\red{$\uparrow$0.02}} & 0.15{\tiny\red{$\uparrow$0.11}} \\
SEED-X~\cite{ge2024seed} & 0.49{\tiny\red{$\uparrow$0.17}} & 0.97{\tiny\red{$\uparrow$0.26}} & 0.58{\tiny\red{$\uparrow$0.24}} & 0.26{\tiny\red{$\uparrow$0.05}} & 0.80{\tiny\red{$\uparrow$0.22}} & 0.19{\tiny\red{$\uparrow$0.12}} & 0.14{\tiny\red{$\uparrow$0.10}} \\
Emu3-Gen~\cite{wang2024emu3} & 0.54{\tiny\red{$\uparrow$0.22}} & 0.98{\tiny\red{$\uparrow$0.27}} & 0.71{\tiny\red{$\uparrow$0.37}} & 0.34{\tiny\red{$\uparrow$0.13}} & 0.81{\tiny\red{$\uparrow$0.23}} & 0.17{\tiny\red{$\uparrow$0.10}} & 0.21{\tiny\red{$\uparrow$0.17}} \\
Janus~\cite{wu2025janus} & 0.61{\tiny\red{$\uparrow$0.29}} & 0.97{\tiny\red{$\uparrow$0.26}} & 0.68{\tiny\red{$\uparrow$0.34}} & 0.30{\tiny\red{$\uparrow$0.09}} & 0.84{\tiny\red{$\uparrow$0.26}} & 0.46{\tiny\red{$\uparrow$0.39}} & 0.42{\tiny\red{$\uparrow$0.38}} \\
JanusFlow~\cite{ma2025janusflow} & 0.63{\tiny\red{$\uparrow$0.31}} & 0.97{\tiny\red{$\uparrow$0.26}} & 0.59{\tiny\red{$\uparrow$0.25}} & 0.45{\tiny\red{$\uparrow$0.24}} & 0.83{\tiny\red{$\uparrow$0.25}} & 0.53{\tiny\red{$\uparrow$0.46}} & 0.42{\tiny\red{$\uparrow$0.38}} \\
Janus-Pro-7B~\cite{chen2025janus} & 0.80{\tiny\red{$\uparrow$0.48}} & 0.99{\tiny\red{$\uparrow$0.28}} & 0.89{\tiny\red{$\uparrow$0.55}} & 0.59{\tiny\red{$\uparrow$0.38}} & 0.90{\tiny\red{$\uparrow$0.32}} & 0.79{\tiny\red{$\uparrow$0.72}} & 0.66{\tiny\red{$\uparrow$0.62}} \\
GoT~\cite{fang2025got} & 0.64{\tiny\red{$\uparrow$0.32}} & 0.99{\tiny\red{$\uparrow$0.28}} & 0.69{\tiny\red{$\uparrow$0.35}} & 0.67{\tiny\red{$\uparrow$0.46}} & 0.85{\tiny\red{$\uparrow$0.27}} & 0.34{\tiny\red{$\uparrow$0.27}} & 0.27{\tiny\red{$\uparrow$0.23}} \\
Bagel~\cite{deng2025emerging} & 0.78{\tiny\red{$\uparrow$0.46}} & 0.98{\tiny\red{$\uparrow$0.27}} & 0.94{\tiny\red{$\uparrow$0.60}} & 0.76{\tiny\red{$\uparrow$0.55}} & 0.91{\tiny\red{$\uparrow$0.33}} & 0.69{\tiny\red{$\uparrow$0.62}} & 0.70{\tiny\red{$\uparrow$0.66}} \\
GPT-Image-1~\cite{gpt_image_1} & \blueboxtext{0.84}{\tiny\red{$\uparrow$0.52}} & \blueboxtext{0.99}{\tiny\red{$\uparrow$0.28}} & \blueboxtext{0.92}{\tiny\red{$\uparrow$0.58}} & \blueboxtext{0.85}{\tiny\red{$\uparrow$0.64}} & \blueboxtext{0.92}{\tiny\red{$\uparrow$0.34}} & \blueboxtext{0.75}{\tiny\red{$\uparrow$0.68}} & \blueboxtext{0.61}{\tiny\red{$\uparrow$0.57}} \\

\midrule
\multicolumn{8}{l}{\textit{Collaborative AI Systems}} \\
ComfyAgent~\cite{xue2025comfybench} & 0.32 & 0.69 & 0.30 & 0.33 & 0.50 & 0.04 & 0.04 \\
ComfyMind~\cite{guo2026comfymind} (reproduced) & 0.90{\tiny\red{$\uparrow$0.58}} & 1.00{\tiny\red{$\uparrow$0.31}} & 1.00{\tiny\red{$\uparrow$0.70}} & 0.96{\tiny\red{$\uparrow$0.63}} & 0.97{\tiny\red{$\uparrow$0.47}} & 0.63{\tiny\red{$\uparrow$0.59}} & 0.81{\tiny\red{$\uparrow$0.77}} \\
\textbf{SymbOmni (Ours)} & \blueboxtext{0.98}{\tiny\red{$\uparrow$0.66}} & \blueboxtext{1.00}{\tiny\red{$\uparrow$0.31}} & \blueboxtext{1.00}{\tiny\red{$\uparrow$0.70}} & \blueboxtext{0.99}{\tiny\red{$\uparrow$0.66}} & \blueboxtext{0.98}{\tiny\red{$\uparrow$0.48}} & \blueboxtext{0.97}{\tiny\red{$\uparrow$0.93}} & \blueboxtext{0.95}{\tiny\red{$\uparrow$0.91}} \\

\Xhline{1.2pt}
\end{tabular}
}
\caption{Evaluation of T2I generation on GenEval~\cite{ghosh2023geneval}. Obj.: Object. Attr.: Attribute. We highlight the best results with colored boxes.}
\label{tab:geneval}
\end{table}

\paragraph{GenEval: Text-to-Image Generation.}
As shown in Table~\ref{tab:geneval}, SymbOmni achieves a state-of-the-art overall score of 0.98, substantially outperforming methods based on frozen text encoders (best: SD3-Medium, 0.74), LLM/MLLM-enhanced approaches (best: GPT-Image-1, 0.84), and collaborative AI systems (best: ComfyMind, 0.90). Its advantage is particularly pronounced on challenging compositional tasks, where it attains near-perfect scores of 0.97 for Position and 0.95 for Attribute Binding. These results underscore the effectiveness of symbolic concept learning in enabling robust and precise compositional reasoning.

\paragraph{Additional Benchmarks and Ablations.}
We further evaluate our method on GenEval2~\cite{kamath2025geneval} and Kris Bench~\cite{wu2026kris}, and include additional ablation studies in the supplementary material to assess the generalization of our method across diverse benchmarks and evaluation settings.

\subsection{Ablation Study}

\begin{table}[t]
\centering
\scriptsize
\setlength{\abovecaptionskip}{6pt}
\setlength{\belowcaptionskip}{-3pt}
\renewcommand{\arraystretch}{1.2}
\renewcommand\tabcolsep{4pt} 
\begin{tabular}{l|ccc|c}
\Xhline{1.2pt}
\rowcolor{CadetBlue!20}
\multirow{2}{*}{\textbf{Method}} & \multicolumn{3}{c|}{\textbf{Avg Tokens/Case}} & \textbf{Avg/Case} \\
\cmidrule(lr){2-4} \cmidrule(lr){5-5}
 & Input{\red{$\downarrow$}} & Output{\red{$\downarrow$}} & Total{\red{$\downarrow$}} & \textbf{Requests}{\red{$\downarrow$}} \\
\Xhline{1.2pt}

\textbf{Ours (w/o EXP)} &
18,556.1 &
3,171.0 &
21,727.1 &
9.85 \\

\textbf{Ours} &
\blueboxtext{12,463.4}{\scriptsize\red{$\downarrow$32.8\%}} &
\blueboxtext{1,898.0}{\scriptsize\red{$\downarrow$40.1\%}} &
\blueboxtext{14,361.4}{\scriptsize\red{$\downarrow$33.9\%}} &
\blueboxtext{7.25}{\scriptsize\red{$\downarrow$26.4\%}} \\

\Xhline{1.2pt}
\end{tabular}%
\caption{Performance and cost statistics on ReasonEdit benchmark. We highlight the best results with colored boxes.}
\label{tab:ablation_performance}
\end{table}

We ablate the symbolic concept memory by removing the memory retrieval module (without EXP). As shown in Table~\ref{tab:ablation_performance}, removing memory retrieval consistently degrades performance across all evaluation metrics~\cite{lewis2020retrieval,brown2020language,wei2022chain}. More notably, the memory module reduces total token consumption by 33.9\%, including reductions of 32.8\% in input tokens and 40.1\% in output tokens, while decreasing the number of API requests per case by 26.4\%. By reusing accumulated experience, the system can identify effective strategies earlier, thereby avoiding failed attempts and reducing iterative replanning~\cite{guo2025deepseek,jaech2024openai,de2024system}. These findings establish symbolic concept memory as a key mechanism for experience reuse, enabling the continual accumulation of task-level knowledge beyond what can be achieved through in-context learning or parametric recall alone~\cite{hu2025automated,shang2025agentsquare,cemri2026multi}.

\subsection{Offline Mode Performance Evaluation}

To reduce inference cost and alleviate the context burden from lengthy workflow descriptions, we design an offline evaluation where the agent operates solely on retrieved symbolic concepts without access to external workflow descriptions.

We evaluate SymbOmni on the 200-task ComfyBench benchmark~\cite{xue2025comfybench}, comprising 100 vanilla, 60 complex, and 40 creative tasks. The tasks are partitioned into 10 groups of 20, with even-indexed tasks in each group assigned to the online set and odd-indexed tasks to the offline set, yielding 100 tasks per split while maintaining comparable difficulty. We shuffle the groups to disrupt the original monotonic ordering while preserving task similarity within each group.
During the online phase, SymbOmni processes the groups sequentially, executing the 10 online tasks in each group in parallel. Experiences are committed to memory only after all tasks in the group have been completed, preventing information leakage within the group. The system then transitions to the offline phase, where it solves the 100 held-out tasks using only the accumulated symbolic concept memory, without access to workflow descriptions. We compare against ComfyMind~\cite{guo2026comfymind}, which retains full access to workflow descriptions but does not maintain an experience memory. This comparison directly tests whether learned symbolic concepts can substitute for explicit documentation. Our analysis focuses on the more challenging Complex category, with additional protocol details provided in the supplementary material.

As shown in Table~\ref{tab:offline_performance}, our offline approach achieves a resolve rate of 70.0\%, surpassing the documentation-dependent baseline. More importantly, it reduces the average number of attempts by 28.4\% across all tasks and by 25.6\% among successfully resolved tasks. These results show that symbolic concept memory not only compensates for the absence of explicit workflow documentation, but also improves problem-solving efficiency through experience-driven optimization. This highlights its practical value in lowering inference costs and mitigating the context overhead associated with lengthy workflow descriptions~\cite{wu2023autogen,xie2023openagents,autogpt}.

\subsection{User Study}
\label{sec:user_study}

Using the generation results from the 38 cases presented in Section~\ref{sec:qualitative_experiment}, we conduct a large-scale user preference study to compare the evaluated methods. We employ a pairwise blind evaluation protocol, in which SymbOmni was compared with each of the four other methods across all 38 cases, resulting in 152 comparison questions. In total, 63 participants contributed 1,197 preference judgments, assessing each pair along four dimensions: semantic consistency, visual quality, compositional accuracy, and overall preference. Additional details of the study design and evaluation protocol are provided in the supplementary material.

\begin{table}[t!]
\centering
\scriptsize
\setlength{\abovecaptionskip}{6pt}
\setlength{\belowcaptionskip}{6pt}
\renewcommand{\arraystretch}{1.2}
\renewcommand\tabcolsep{6pt}
\begin{tabular}{l|ccc}
\Xhline{1.2pt}
\rowcolor{CadetBlue!20}
\textbf{Method} & \textbf{Resolved}{\red{$\uparrow$}} & \textbf{Avg Tries}{\red{$\downarrow$}} & \textbf{Avg Tries (Resolved){\red{$\downarrow$}}} \\
\Xhline{1.2pt}
ComfyMind (w/ description) & 66.7\% & 2.000 & 1.600 \\
\textbf{Ours (Offline)} & \blueboxtext{70.0\%} {\tiny\red{$\uparrow$3.3\%}}& \blueboxtext{1.433}{\tiny\red{$\downarrow$0.567}} & \blueboxtext{1.190}{\tiny\red    {$\downarrow$0.410}} \\
\Xhline{1.2pt}
\end{tabular}
\caption{Offline mode performance on the Complex offline split ($N=30$).}
\label{tab:offline_performance}
\end{table}

\begin{table}[t!]
    \centering
    \scriptsize
    \setlength{\abovecaptionskip}{6pt}
\setlength{\belowcaptionskip}{5pt}
\renewcommand{\arraystretch}{1.2}
    \renewcommand\tabcolsep{4pt}
    \begin{tabular}{l|ccccc}
        \Xhline{1.2pt}
        \rowcolor{CadetBlue!20}
        \textbf{Method} & \textbf{Van.\ Pass} & \textbf{Van.\ Resolve} & \textbf{Cplx.\ Resolve} & \textbf{Crtv.\ Resolve} & \textbf{Overall} \\
        \Xhline{1.2pt}
        ComfyMind~\cite{guo2026comfymind} & 100.0\% & 84.0\% & 55.0\% & 42.5\% & 67.0\% \\
        \textbf{SymbOmni} & 100.0\% & \textbf{89.0\%} & \textbf{66.7\%} & \textbf{52.5\%} & \textbf{75.0\%} \\
        \midrule
        Improvement & +0.0\% & +5.0\% & +11.7\% & +10.0\% & +8.0\% \\
        \Xhline{1.2pt}
    \end{tabular}
    \caption{Generalization on out-of-domain workflows and ComfyBench.}
    \label{tab:workflow_ablation}
\end{table}

\begin{figure}[t!]
\centering
\setlength{\abovecaptionskip}{6pt}
\setlength{\belowcaptionskip}{-4pt}
\definecolor{myblue}{RGB}{50, 115, 180}
\definecolor{mygray}{RGB}{190, 190, 190}
\resizebox{0.91\linewidth}{!}{
\begin{tikzpicture}
    \begin{axis}[
        xbar stacked,
        bar width=22pt,
        width=10cm,
        height=5.5cm,
        xmin=0, xmax=100,
        ymin=0.5, ymax=4.5,
        clip=false,
        axis on top,
        axis line style={draw=black, line width=0.8pt},
        tick style={draw=black, major tick length=3pt},
        ytick={1,2,3,4},
        yticklabels={Nano Banana~\cite{comanici2025gemini}, ComfyMind~\cite{guo2026comfymind}, GPT-Image-1~\cite{gpt_image_1}, BAGEL~\cite{deng2025emerging}},
        yticklabel style={
            anchor=west,
            xshift=2pt,
            font=\small
        },
        ytick pos=right,
        xlabel={Preference Rate (\%)},
        xtick={0, 20, 40, 60, 80, 100},
        xticklabel style={font=\small},
        xlabel style={font=\small, yshift=-1ex},
        legend style={draw=none, fill=none},
        nodes near coords,
        nodes near coords align={center},
        every node near coord/.append style={
            font=\small,
            text=white,
            /pgf/number format/fixed,
            /pgf/number format/precision=1
        }
    ]

    \draw [dashed, thick, black!70] (axis cs:50, 0.5) -- (axis cs:50, 4.6);

    \node[
        anchor=south,
        font=\scriptsize,
        text=black!70,
        inner sep=1pt,
        yshift=2pt
    ] at (axis cs:50, 4.6) {50\% Threshold};

    \addplot+[myblue, fill=myblue, draw=none, text=white] coordinates {
        (59.2,1) (65.9,2) (68.4,3) (73.9,4)
    };

    \addplot+[mygray, fill=mygray, draw=none, text=white] coordinates {
        (40.8,1) (34.1,2) (31.6,3) (26.1,4)
    };

    \node[anchor=east, font=\small, xshift=-4pt] at (axis cs:0,1) {SymbOmni};
    \node[anchor=east, font=\small, xshift=-4pt] at (axis cs:0,2) {SymbOmni};
    \node[anchor=east, font=\small, xshift=-4pt] at (axis cs:0,3) {SymbOmni};
    \node[anchor=east, font=\small, xshift=-4pt] at (axis cs:0,4) {SymbOmni};

    \end{axis}
\end{tikzpicture}
}
\caption{Pairwise preference rates of SymbOmni vs. other methods.}
\label{fig:user_preference}
\end{figure}
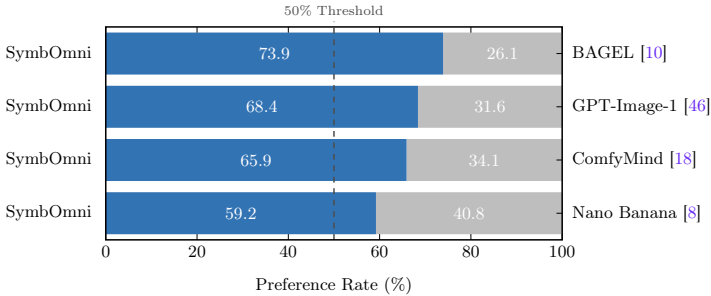

As shown in Figure~\ref{fig:user_preference}, SymbOmni is consistently preferred across all pairwise comparisons. It achieves a preference rate of 59.2\% over Nano Banana, with the margin increasing to 65.9\% against ComfyMind and 68.4\% against GPT-Image-1. These results highlight the advantages of symbolic concept learning over heuristic search and purely end-to-end generation paradigms.
The largest preference margin is observed against BAGEL, where SymbOmni attains 73.9\%, demonstrating the effectiveness of combining agentic workflow decomposition with symbolic memory for compositional reasoning and creative generation.

\subsection{Out-of-Domain Generalization}
\label{sec:main_generalization}

To evaluate whether the learned symbolic concepts generalize beyond the source distribution, we construct an external concept library from 147 in-the-wild workflows spanning diverse domains, following the data collection protocol of ComfyGPT~\cite{huang2025comfygpt}. These workflows encompass a broad range of visual generation and editing paradigms and are entirely disjoint from the original training and evaluation pipeline. In practice, SymbOmni performs symbolic concept learning on this external workflow set and is subsequently evaluated on ComfyBench~\cite{xue2025comfybench}, which can test whether concepts acquired from substantially different workflow distributions can transfer to structured benchmark tasks or not.

As shown in Table~\ref{tab:workflow_ablation}, SymbOmni achieves consistent improvements across all difficulty levels. The overall resolve rate improves from 67.0\% to 75.0\% (a 8.0\% gain), with the most pronounced gains in harder categories, including Complex +11.7\% (55.0\% $\to$ 66.7\%) and Creative +10.0\% (42.5\% $\to$ 52.5\%). 
These results demonstrate that symbolic concepts extracted from diverse, in-the-wild workflows are able to transfer effectively to structured benchmark tasks, confirming the ability of SymbOmni's concept learning mechanism to generalize beyond the workflow distributions encountered during evolution.

We further evaluate SymbOmni across a diverse set of additional benchmarks, and SymbOmni can still consistently achieve state-of-the-art or competitive performance. Our cross-modal generalization analysis also demonstrates that the learned symbolic concepts transfer effectively across different combinations of modalities. Complete results are provided in the supplementary material.

\section{Concluding Remarks}
\label{sec:conclusion}

We introduce \textbf{\symbomni}, a cognitive architecture for multimodal creation that addresses the perpetual novice problem through Symbolic Concept Learning. By abstracting experience into reusable concepts through an Induction-Transduction cycle guided by verbalized backpropagation, SymbOmni enables continual learning and self-improvement without parameter updates. Extensive experiments demonstrate SymbOmni's strong performance, improved problem-solving efficiency, and continuous gains from accumulated experience, including over 40\% reduction in token consumption. Its dual-feedback mechanism further enables the system to learn from both successful and failed executions, providing a principled foundation for adaptive AI systems that continuously refine their behavior. By integrating symbolic reasoning with experiential learning, this work offers a promising path toward scalable and sustainable AI evolution.

%% file: sec/X_suppl.tex
\section{Appendix Overview}
\label{sec:overview}

\begin{itemize}
\item \textbf{Sec.~\ref{sec:plugin}} introduces the ComfyUI-SymbOmni plugin for practical deployment and community adoption.

\item \textbf{Sec.~\ref{sec:qualitative_comparison}} presents the complete qualitative comparison across 38 test cases covering text-to-image synthesis, complex compositional generation, creative content creation, image editing tasks, and text-to-video generation.

\item \textbf{Sec.~\ref{sec:user_study_details}} details the user study design, including the pairwise blind comparison protocol and evaluation dimensions across 1,197 responses.

\item \textbf{Sec.~\ref{sec:additional_experiments}} provides additional quantitative experiments, including comprehensive evaluation on the offline mode performance assessment and LLM ablation study.

\item \textbf{Sec.~\ref{sec:generalization}} presents comprehensive generalization analysis, including detailed experimental protocol for the out-of-domain workflow generalization experiment, cross-benchmark evaluation across four diverse benchmarks, and cross-modal generalization assessment.

\item \textbf{Sec.~\ref{sec:scalability}} analyzes the scalability of the Symbolic Concept Box, including retrieval accuracy when merging concept libraries from different domains.

\item \textbf{Sec.~\ref{sec:self_correction}} discusses the self-correction mechanism in symbolic concept learning and how the system addresses memory pollution through continuous optimization.

\item \textbf{Sec.~\ref{sec:prompt_set}} contains the complete prompt set and evaluation protocol details used in the SymbOmni system.
\end{itemize}

\newpage
\section{ComfyUI-SymbOmni Plugin}
\label{sec:plugin}

To facilitate practical deployment and community adoption, we release a ComfyUI-SymbOmni plugin based on ComfyUI-Copilot~\cite{xu2025comfyui} within the ComfyUI ecosystem~\cite{comfyui}. We remove the closed-source backend dependencies in the original Copilot; all planning, step execution, workflow optimization, and concept learning management are performed locally with open-source components, powered by SymbOmni. The plugin integrates symbolic concept learning capabilities into the ComfyUI GUI, enabling users to interactively manage symbolic concepts, personal workflows, workflow experiences and optimize workflows through an intuitive node-based interface.

\section{Complete Qualitative Comparison}
\label{sec:qualitative_comparison}

This section presents visual generation results across 38 carefully designed test cases, evaluated using five different methods. These cases systematically cover major application scenarios in visual generation, including simple text-to-image synthesis, complex compositional generation, creative content creation, image editing tasks, and text-to-video generation. We present 31 representative cases in the following figures; the 38 image generation cases also serve as the evaluation set for the user study (Sec.~\ref{sec:user_study_details}).

\subsection{Test Case Design}

We constructed 38 representative cases that systematically span the primary challenges and application scenarios in visual generation. The cases are organized into four major categories: simple text-to-image tasks (8 cases) covering basic object-scene combinations such as ``Generate an image of a cat sitting on a windowsill looking outside'' and ``Generate an image of a futuristic factory filled with robots assembling machines''; complex compositional generation (6 cases) featuring multi-object spatial arrangements and attribute binding challenges, exemplified by ``Breaking Bad movie poster with Walter White in white vest'' and geometric configurations with precise color-position specifications; Creative content generation (4 cases) explores diverse artistic creations, such as a four-panel comic strip depicting a man discovering it's Sunday and staying in bed, and designing a visually striking book cover for ``A Song of Ice and Fire'' that incorporates a dragon, a snowy landscape, and a sword; and image editing tasks (20 cases) encompassing color transformation(1 case), style transfer(3 cases), object removal(4 cases), object insertion(3 cases), image matting(4 cases), reasoning editing(3 cases),and object manipulation(2 cases) such as converting ``red car to green car with canvas extension'' and ``Donald Trump photo to clay sculpture style.''

\subsection{Comparison Methods}

For each case, we generated results using five distinct approaches. SymbOmni, our proposed method, employs Gemini 2.5 Flash~\cite{comanici2025gemini} as the planning engine with symbolic concept retrieval and self-evolution capabilities enabled. ComfyMind~\cite{guo2026comfymind} serves as the baseline agent system, utilizing the same reasoning engine but relying on heuristic search without symbolic concept learning. BAGEL~\cite{deng2025emerging} represents unified multimodal models with end-to-end generation from text to image and from image to image. Nano Banana~\cite{comanici2025gemini} and GPT-Image-1~\cite{gpt_image_1} provide state-of-the-art closed-source unified model performance for comparison.

\begin{figure}[t!]
    \centering
    \includegraphics[width=0.95\linewidth]{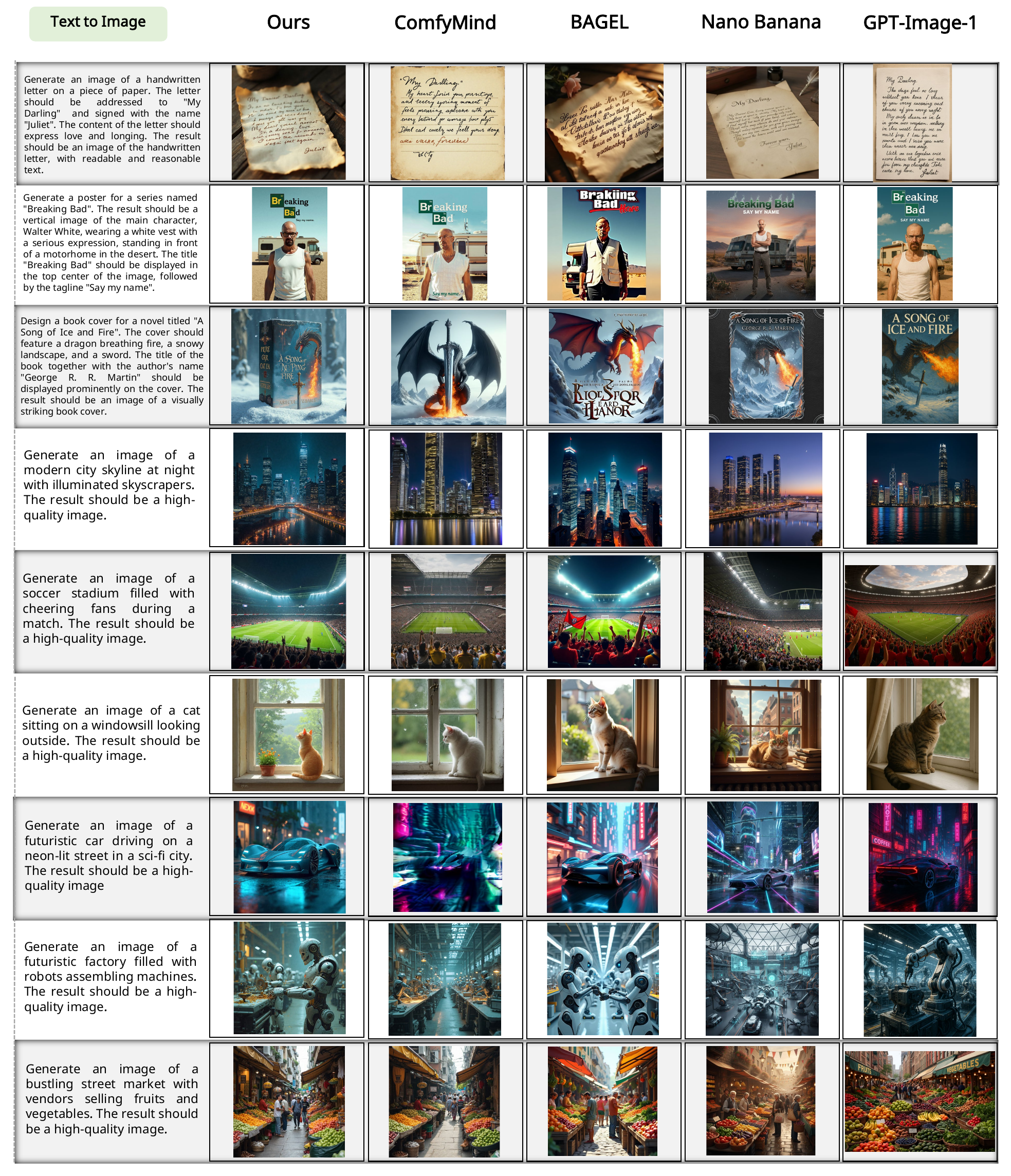}
    \caption{Qualitative comparison results for text-to-image generation.  }
    \label{fig:qualitative_results_t2i}
\end{figure}

\begin{figure}[t!]
    \centering
    \includegraphics[width=0.95\linewidth]{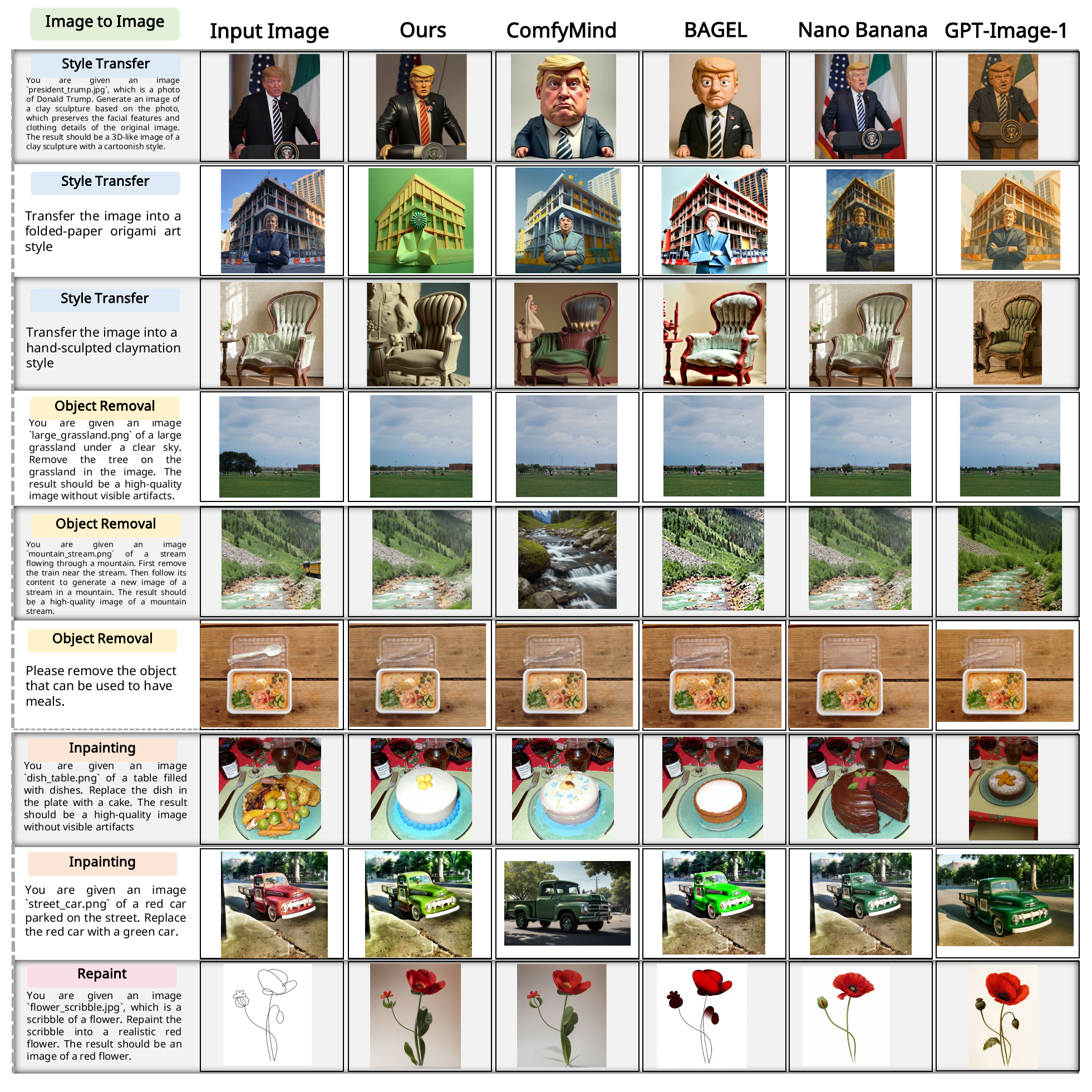}
    \caption{Qualitative comparison results for image-to-image generation.}
    \label{fig:qualitative_results_i2i_1}
\end{figure}

\begin{figure}[t!]
    \centering
    \includegraphics[width=0.95\linewidth]{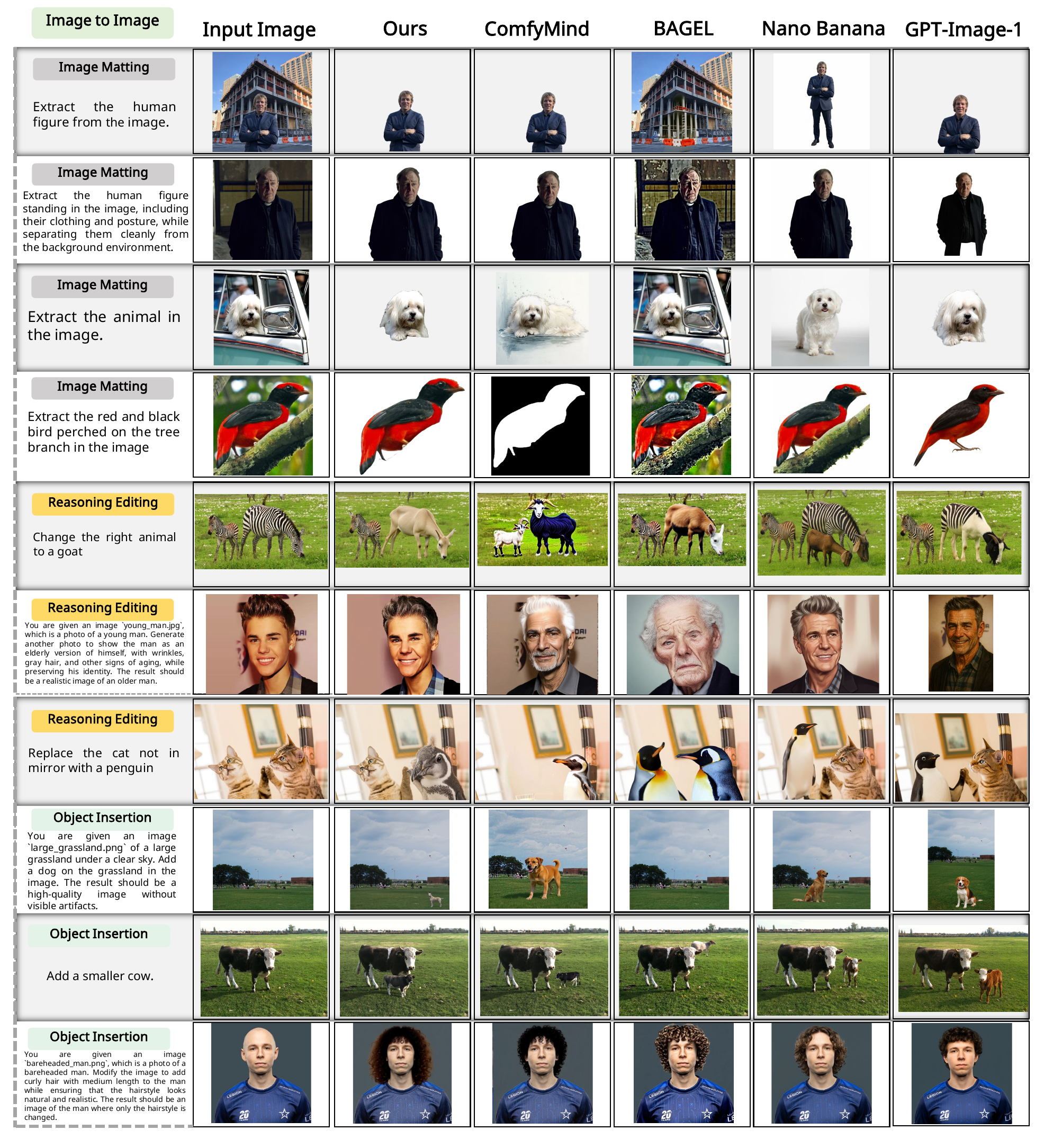}
    \caption{Qualitative comparison results for image-to-image generation.}
    \label{fig:qualitative_results_i2i_2}
\end{figure}

\begin{figure}[h!]
    \centering
    \vspace{-5mm}
    \includegraphics[width=0.83\linewidth]{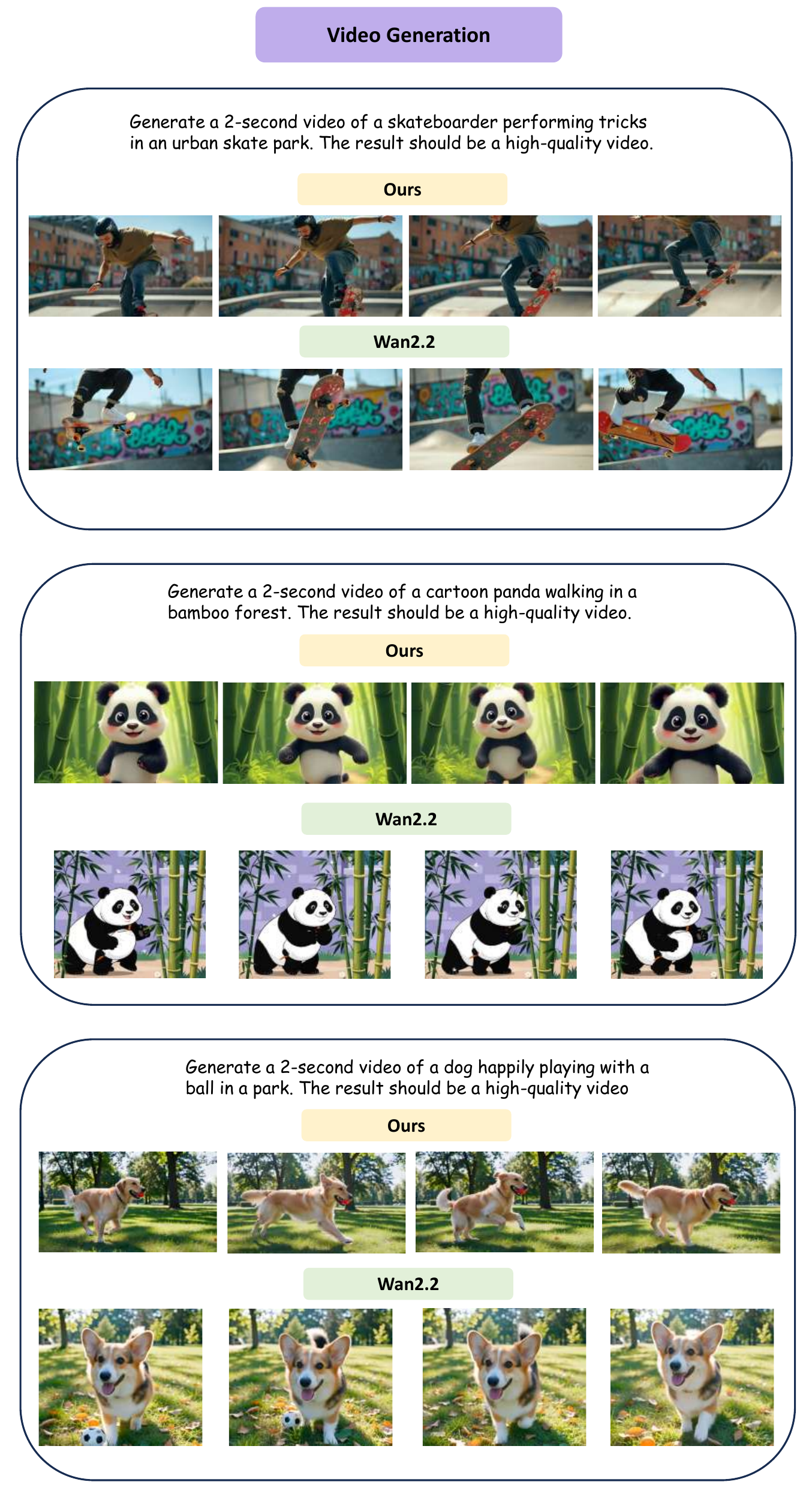}
    \caption{Qualitative comparison results for Text-to-Video generation.}
    \label{fig:qualitative_results_video}
    \vspace{-5mm}
\end{figure}

\section{User Study Details}
\label{sec:user_study_details}

This section provides the detailed protocol for the user study summarized in the main text (Section~\ref{sec:user_study}).

\subsection{Study Design}

We adopted a pairwise blind comparison methodology. From the five methods' image generation results across 38 cases, SymbOmni was compared one-on-one against each of the four alternative methods, yielding a total of 152 comparison questions (38 cases $\times$ 4 comparisons).

We recruited 63 volunteers to participate in the evaluation. To ensure comprehensive coverage while maintaining evaluation quality, each volunteer was assigned to evaluate 19 image pairs randomly selected from the 152 comparison questions. The blind testing protocol ensured complete randomization of image display order, with no method identification visible to participants who only saw labels such as ``Image A'' and ``Image B''. Text prompts and input image were clearly displayed above each image pair, and participants remained unaware of any method information until completing all evaluations.

\subsection{Evaluation Dimensions}

Through this evaluation process, we collected a total of 1,197 preference responses (63 volunteers $\times$ 19 pairs). Participants evaluated each image pair across four comprehensive dimensions:

\vspace{1.5mm}

\begin{itemize}[leftmargin=*, nosep]
    \item \textbf{Semantic Consistency}: measured how well the generated result matched the text prompt's intent.
    \item \textbf{Visual Quality}: assessed overall image quality including clarity, coherence, and realism.
    \item \textbf{Compositional Accuracy}: assessed whether the generated result presented a coherent and faithful composition of the
  requested elements, attributes, and spatial relations.
    \item \textbf{Overall Preference}: represented participants' overall judgment considering all factors.
\end{itemize}

\section{Additional Quantitative Experiments}
\label{sec:additional_experiments}

\subsection{Offline Mode Performance Evaluation Details}

This section provides the detailed experimental protocol for the offline mode evaluation summarized in the main paper. We elaborate on the data partitioning strategy, execution synchronization, and design rationale below.

\subsubsection{Detailed Experimental Protocol}

To fairly evaluate the performance of online and offline modes, we carefully designed a controlled experimental protocol. ComfyBench~\cite{xue2025comfybench} contains 200 tasks with three difficulty levels: 100 vanilla, 60 complex, and 40 creative tasks. These tasks are arranged in ascending order of difficulty; adjacent tasks may involve overlapping skill requirements (e.g., both requiring inpainting or ControlNet), but differ in their specific instructions, object compositions, and expected outputs, so that solving one task does not directly transfer a ready-made solution to its neighbor.

We first partition the 200 tasks into 10 consecutive groups of 20 tasks each, preserving the original ordering. Within each group, we perform an even-odd split based on task indices: even-indexed tasks are assigned to the online set, while odd-indexed tasks form the offline set. This stratified splitting ensures that both sets maintain comparable difficulty distributions—each group contributes 10 online and 10 offline tasks, corresponding to approximately 5 vanilla, 3 complex, and 2 creative tasks per set per group.

To mitigate potential biases from the original task ordering while maintaining experimental reproducibility, we randomly shuffle the order of these 10 groups (rather than shuffling individual tasks). This group-level randomization serves two critical purposes: (1) it breaks the monotonic difficulty progression across the entire benchmark, preventing the online mode from systematically encountering easier or harder tasks at specific learning stages; (2) it preserves the within-group difficulty balance and enables controlled parallel execution, as detailed below.

During the online phase, SymbOmni processes the 10 groups sequentially according to the shuffled order. Within each group, all 10 online tasks are executed in parallel using 10 worker processes. Crucially, experience generated within a group is only summarized and uploaded to the experience repository after all tasks in that group complete. This group-wise synchronization ensures that: (1) tasks within the same group cannot leverage each other's experiences, providing a conservative lower bound on online learning performance; (2) the learning trajectory is deterministic and reproducible given a fixed group ordering, independent of system-level factors such as task scheduling latency or hardware variability.

After completing all 100 online tasks across the 10 groups, the system enters offline mode. Using the accumulated experience repository (but without access to original workflow descriptions), SymbOmni attempts to solve the 100 offline tasks using the same group-sequential execution strategy. This experimental design allows us to analyze learning curves across different experience accumulation stages (e.g., comparing performance of group \(i\) with access to experiences from groups \(1\) to \(i-1\)) and to quantify the performance gap between online learning (progressive experience accumulation during task execution) and offline generalization (leveraging a complete experience repository).

\subsubsection{Results and Analysis}

The experiment comprises two phases: (1) an online phase where the agent completes 100 ComfyBench~\cite{xue2025comfybench} tasks to build its experience memory; (2) an offline phase where it tackles another 100 tasks using only accumulated symbolic knowledge. We compare our offline approach against ComfyMind~\cite{guo2026comfymind}, which has full access to workflow descriptions but lacks experience memory. This comparison directly tests whether learned symbolic concepts can effectively substitute for explicit documentation. We focus our analysis on the challenging \textbf{Complex} task category.

\begin{table}[t!]
\centering
\renewcommand\tabcolsep{6pt}
\resizebox{\linewidth}{!}{
\begin{tabular}{l|ccc}
\Xhline{1.2pt}
\rowcolor{CadetBlue!20}
\textbf{Method} & \textbf{Resolved}{\red{$\uparrow$}} & \textbf{Avg Tries}{\red{$\downarrow$}} & \textbf{Avg Tries (Resolved){\red{$\downarrow$}}} \\
\Xhline{1.2pt}
ComfyMind (w/ description) & 66.7\% & 2.000 & 1.600 \\
\textbf{Ours (Offline)} & \blueboxtext{70.0\%} {\tiny\red{$\uparrow$3.3\%}}& \blueboxtext{1.433}{\tiny\red{$\downarrow$0.567}} & \blueboxtext{1.190}{\tiny\red{$\downarrow$0.410}} \\
\Xhline{1.2pt}
\end{tabular}
}
\vspace{2mm}
\caption{Offline mode performance on \textbf{Complex} tasks.}
\label{tab:offline_performance_sup}
\end{table}

As shown in Table~\ref{tab:offline_performance_sup}, our offline approach achieves a 70.0\% resolve rate, outperforming the documentation-dependent baseline. More significantly, it reduces the average attempt count by 28.4\% and by 25.6\% for successfully resolved tasks. These results demonstrate that symbolic concept memory not only compensates for the absence of explicit documentation but enables more efficient problem-solving through experience-based optimization, highlighting its practical value in deployment scenarios with limited external resources.

\subsection{LLM Ablation Study}

To validate the adaptability of the SymbOmni framework across different LLM backbones, we conducted systematic ablation experiments comparing closed-source and open-source language models.

\subsubsection{Experimental Design}

The ablation study evaluates SymbOmni's performance when powered by different reasoning engines. We compared Gemini 2.5 Flash~\cite{comanici2025gemini}, a state-of-the-art cost‑effective closed-source model, against an open-source combination consisting of Qwen3-VL-235B-A22B-Instruct~\cite{yang2025qwen3} for visual understanding and overall task planning, paired with DeepSeek-V3.2-Exp~\cite{guo2025deepseek} for text completion tasks. The evaluation was conducted on ComfyBench~\cite{xue2025comfybench} to assess workflow construction capabilities across vanilla, complex, and creative task categories.

\subsubsection{Results and Analysis}

\begin{table}[t!]
    \centering
    \scriptsize
    \renewcommand\tabcolsep{5pt}
    \resizebox{\linewidth}{!}{
    \begin{tabular}{l|cccccccc}
        \Xhline{1.2pt}
        \rowcolor{CadetBlue!20}
        \multirow{2}{*}{\textbf{LLM Backend}} & \multicolumn{2}{c}{\textbf{Vanilla}} & \multicolumn{2}{c}{\textbf{Complex}} & \multicolumn{2}{c}{\textbf{Creative}} & \multicolumn{2}{c}{\textbf{Total}} \\
        \cmidrule(lr){2-3} \cmidrule(lr){4-5} \cmidrule(lr){6-7} \cmidrule(lr){8-9}
        & \%Pass{\red{$\uparrow$}} & \%Resolve{\red{$\uparrow$}} & \%Pass{\red{$\uparrow$}} & \%Resolve{\red{$\uparrow$}} & \%Pass{\red{$\uparrow$}} & \%Resolve{\red{$\uparrow$}} & \%Pass{\red{$\uparrow$}} & \%Resolve{\red{$\uparrow$}} \\
        \Xhline{1.2pt}
        Gemini 2.5 Flash & \blueboxtext{100.0} & \blueboxtext{95.0} & \blueboxtext{100.0} & \blueboxtext{83.3} & \blueboxtext{100.0} & \blueboxtext{67.5} & \blueboxtext{100.0} & \blueboxtext{86.0} \\
        Qwen3VL+DeepSeek & 100.0{\tiny\red{0.0}} & 94.0{\tiny\red{$\downarrow$1.0}} & 100.0{\tiny\red{0.0}} & 80.0{\tiny\red{$\downarrow$3.3}} & 100{\tiny\red{0.0}} & 55.0{\tiny\red{$\downarrow$12.5}} & 100{\tiny\red{0.0}} & 82.0{\tiny\red{$\downarrow$4.0}} \\
        \Xhline{1.2pt}
    \end{tabular}
    }
    \vspace{2mm}
    \caption{LLM Ablation Study on ComfyBench~\cite{xue2025comfybench}: Closed-source vs. Open-source Reasoning Engines}
    \label{tab:llm_ablation}
\end{table}

The ablation study reveals several important findings regarding LLM backbone selection for the SymbOmni framework. While the open-source combination achieves competitive performance in vanilla tasks (94.0\% resolve rate), a notable gap emerges in complex scenarios where Gemini 2.5 Flash~\cite{comanici2025gemini} attains 83.3\% compared to 80.0\% for the open-source variant, indicating an 3.3 percentage point advantage. The disparity becomes more pronounced in creative tasks, where Gemini 2.5 Flash achieves 67.5\% resolve rate versus 55.0\% for the open-source combination, suggesting that closed-source models currently maintain an edge in handling highly abstract and creative generation tasks. Despite these differences, the open-source combination demonstrates the framework's fundamental viability with alternative LLM backends, achieving 82.0\% overall resolve rate compared to 86.0\% for the closed-source configuration. This 4.0 percentage point gap, while meaningful, confirms that SymbOmni's symbolic concept learning mechanism remains effective across different reasoning engines, establishing a foundation for future improvements as open-source models continue to advance.

\section{Generalization Analysis}
\label{sec:generalization}

A central question for any learning-based system is whether the acquired knowledge generalizes beyond the training distribution. In this section, we provide comprehensive evidence that SymbOmni's symbolic concept learning mechanism generalizes effectively across out-of-domain workflows, diverse benchmarks, and multiple modalities.

\subsection{Out-of-Domain Workflow Generalization: Experimental Details}

This section provides the detailed experimental protocol for the out-of-domain generalization experiment summarized in the main paper (Section~\ref{sec:main_generalization}). We elaborate on the workflow collection strategy, concept learning procedure, evaluation protocol, and analysis of the results presented in Table~\ref{tab:workflow_ablation} of the main text.

\subsubsection{Out-of-Domain Workflow Collection}

To construct a concept library that is strictly disjoint from the training and evaluation distributions, we collected 147 in-the-wild workflows from the ComfyUI community ecosystem~\cite{comfyui}, following the out-of-domain collection methodology of ComfyGPT~\cite{huang2025comfygpt}. These workflows were curated from publicly shared community repositories, user forums, and open-source workflow libraries, with no overlap with the ComfyBench~\cite{xue2025comfybench} benchmark tasks or the workflows used during SymbOmni's original training phase.

The 147 collected workflows span a broad range of visual generation and editing paradigms, including but not limited to: (1) \textbf{Artistic style transfer} workflows covering oil painting, watercolor, pixel art, and anime stylization; (2) \textbf{Product photography} pipelines with background replacement and lighting adjustment; (3) \textbf{Character design} workflows combining pose estimation, reference-guided generation, and face restoration; (4) \textbf{Architectural visualization} pipelines with ControlNet-based layout conditioning; (5) \textbf{Video generation and editing} workflows involving frame interpolation and temporal consistency modules; and (6) \textbf{Advanced inpainting and outpainting} workflows with segmentation-guided masking and multi-step refinement. This diversity ensures that the concept library captures a wide spectrum of workflow design patterns and node composition strategies that differ substantially from those present in the evaluation benchmark.

\subsubsection{Concept Learning Procedure}

SymbOmni performed symbolic concept learning on the 147 out-of-domain workflows through the standard induction pipeline described in the main paper. Each workflow was processed through the following steps:

\begin{enumerate}[leftmargin=*, nosep]
    \item \textbf{Workflow Parsing}: Each raw ComfyUI workflow JSON was parsed to extract the node graph structure, parameter configurations, and inter-node connections.
    \item \textbf{Semantic Analysis}: The LLM reasoning engine analyzed the workflow to generate a natural language description ($Desc_k$) and identify the core functional purpose and applicable task categories.
    \item \textbf{Symbolic Abstraction}: The concrete workflow was abstracted into a Symbolic Workflow Instruction ($SWI_k$) with parameterized slots, capturing the essential workflow pattern while allowing task-specific adaptation.
    \item \textbf{Concept Consolidation}: Functionally similar concepts were clustered and consolidated to produce non-redundant, high-quality symbolic concepts, following the semantic clustering and consolidation procedure described in the main paper.
\end{enumerate}

The resulting concept library was then integrated into SymbOmni's Symbolic Concept Box (CB), entirely replacing the original concept library to ensure a strict out-of-domain evaluation setting. No concepts from the original training phase were retained during this evaluation.

\subsubsection{Evaluation Protocol}

The evaluation was conducted on the full ComfyBench~\cite{xue2025comfybench} benchmark, which comprises 200 tasks across three difficulty levels: 100 Vanilla, 60 Complex, and 40 Creative tasks. We compared the following configurations:

\begin{itemize}[leftmargin=*, nosep]
    \item \textbf{ComfyMind~\cite{guo2026comfymind}}: The baseline agent system using the same reasoning engine but without symbolic concept learning. ComfyMind relies on heuristic search with access to the full workflow documentation.
    \item \textbf{SymbOmni (Out-of-Domain)}: SymbOmni equipped with the concept library learned exclusively from the 147 in-the-wild workflows, without access to any ComfyBench-specific workflow descriptions or experiences.
\end{itemize}

Both configurations used Gemini 2.5 Flash~\cite{comanici2025gemini} as the planning engine and were evaluated under identical computational conditions. The evaluation metrics include \textbf{Pass Rate} (percentage of tasks producing a syntactically valid workflow) and \textbf{Resolve Rate} (percentage of tasks producing a semantically correct output), consistent with the ComfyBench evaluation protocol~\cite{xue2025comfybench}.

\subsubsection{Results and Analysis}

As reported in Table~\ref{tab:workflow_ablation} of the main paper, SymbOmni achieves consistent and substantial improvements across all difficulty levels when leveraging concepts learned from out-of-domain workflows. The overall resolve rate improves from 67.0\% (ComfyMind~\cite{guo2026comfymind} baseline) to 75.0\%, representing an 8.0 percentage point gain. The most pronounced improvements emerge in the more challenging categories: Complex tasks improve by 11.7 percentage points (55.0\% $\to$ 66.7\%) and Creative tasks by 10.0 percentage points (42.5\% $\to$ 52.5\%).

Several key observations merit further discussion:

\paragraph{Transfer of Abstract Workflow Patterns} The improvement is particularly notable because the 147 in-the-wild workflows were collected independently of ComfyBench~\cite{xue2025comfybench} and cover different task distributions. This indicates that symbolic concept learning captures abstract workflow composition patterns (e.g., ``use ControlNet conditioning for layout-sensitive tasks'' or ``apply iterative refinement for high-fidelity outputs'') that transfer across task domains, rather than memorizing specific workflow instances.

\paragraph{Difficulty-Dependent Gains} The larger improvements on Complex (+11.7\%) and Creative (+10.0\%) tasks compared to Vanilla (+5.0\%) suggest that symbolic concepts are most beneficial when tasks require sophisticated multi-step planning and creative node composition. For simpler Vanilla tasks, the baseline heuristic search is often sufficient, whereas harder tasks benefit substantially from the structured prior knowledge encoded in symbolic concepts.

\paragraph{Knowledge Gap Bridging} SymbOmni achieves these gains \textit{without} access to any ComfyBench~\cite{xue2025comfybench} workflow documentation, while ComfyMind~\cite{guo2026comfymind} has full access to workflow descriptions. This demonstrates that the symbolic concept library effectively substitutes for explicit documentation, encoding equivalent or superior knowledge in a more compact and retrieval-friendly format.

\subsection{Cross-Benchmark Evaluation}
\label{sec:cross_benchmark}

To further validate the breadth of SymbOmni's generalization, we evaluated the framework across four diverse benchmarks that span different aspects of multimodal generation and reasoning. These benchmarks were selected to cover a wide spectrum of capabilities: GenEval2~\cite{kamath2025geneval} for fine-grained text-to-image alignment, WISE~\cite{niu2025wise} for world-knowledge-informed semantic evaluation, ComfyBench~\cite{xue2025comfybench} for autonomous workflow construction, and Kris-Bench~\cite{wu2026kris} for knowledge-rich multimodal reasoning tasks.

\paragraph{Text-to-Image Alignment (GenEval2)} On the GenEval2 benchmark~\cite{kamath2025geneval}, we adopt the Soft-TIFA$_{\mathrm{AM}}$ metric proposed therein. Soft-TIFA extends TIFA~\cite{hu2023tifa} by generating one template-based yes/no question per atomic element (object, attribute, or relation) in the prompt and assigning each answer a \emph{soft} VQA probability rather than a binary label; the subscript AM denotes that the per-image score is the \emph{arithmetic mean} of these per-atom probabilities, thereby measuring atom-level text-image alignment. SymbOmni without experience obtains a Soft-TIFA$_{\mathrm{AM}}$ Overall score of 71.06. Experience-driven evolution alone raises this to 77.34, and further upgrading the backbone to Nano Banana yields 85.66. Compared to the strongest published baselines—Gemini (83.48), Qwen (77.26), and Bagel+CoT (66.56)—the full SymbOmni (w/ Exp.+Banana) configuration outperforms all existing methods, demonstrating that the combination of symbolic concept learning and a stronger generation backbone substantially enhances text-to-image alignment.

\begin{table*}[t!]
    \centering
    \renewcommand{\arraystretch}{1.08}
    \renewcommand\tabcolsep{4pt}
    \resizebox{\linewidth}{!}{
    \begin{tabular}{l|ccccc|c}
        \Xhline{1.2pt}
        \rowcolor{CadetBlue!20}
        \textbf{Method} & \textbf{Object} & \textbf{Attribute} & \textbf{Count} & \textbf{Position} & \textbf{Verb} & \textbf{Overall} \\
        \Xhline{1.2pt}
        \multicolumn{7}{l}{\textit{Stable Diffusion Model Series}} \\
        SD 2.1~\cite{rombach2022high} & 55.1 & 30.4 & 22.3 & 11.7 & 17.8 & 27.46 \\
        SDXL~\cite{podell2024sdxl} & 74.1 & 42.4 & 28.7 & 16.0 & 28.9 & 38.02 \\
        SD3~\cite{esser2024scaling} & 87.0 & 65.5 & 49.9 & 41.0 & 46.7 & 58.02 \\
        SD3.5-large~\cite{esser2024scaling} & 91.6 & 70.3 & 52.2 & 39.5 & 55.6 & 61.84 \\
        \midrule
        \multicolumn{7}{l}{\textit{State-of-the-Art T2I Models}} \\
        Flux.1-dev~\cite{labs2025flux1kontextflowmatching} & 88.4 & 68.3 & 55.6 & 37.0 & 44.4 & 58.74 \\
        Bagel+CoT~\cite{deng2025emerging} & 92.9 & 75.9 & 55.6 & 50.6 & 57.8 & 66.56 \\
        Qwen-Image~\cite{wu2025qwen} & \textbf{99.1} & 85.6 & 70.3 & 60.2 & 71.1 & 77.26 \\
        Gemini 2.5 Flash Image~\cite{comanici2025gemini} & \uline{99.0} & \uline{91.4} & 70.1 & \uline{70.2} & \uline{86.7} & \uline{83.48} \\
        \midrule
        \multicolumn{7}{l}{\textit{Ours}} \\
        SymbOmni (w/o Exp.) & 86.7 & 79.8 & 66.8 & 62.5 & 59.5 & 71.06 \\
        SymbOmni (w/ Exp.) & 95.0 & 83.6 & \uline{74.8} & 68.8 & 64.5 & 77.34 \\
        \textbf{SymbOmni (w/ Exp.+Banana)} & \uline{99.0} & \textbf{93.4} & \textbf{77.3} & \textbf{71.2} & \textbf{87.4} & \textbf{85.66} \\
        \Xhline{1.2pt}
    \end{tabular}
    }
    \vspace{2mm}
    \caption{Detailed GenEval2 per-skill comparison using Soft-TIFA$_{\mathrm{AM}}$~\cite{kamath2025geneval}. Published baselines are copied from the GenEval2 paper. Overall is the arithmetic mean of all five skill scores.}
    \label{tab:geneval2_skill}
\end{table*}

Table~\ref{tab:geneval2_skill} provides a detailed per-skill analysis on the GenEval2 benchmark~\cite{kamath2025geneval}. Experience-driven evolution (w/o Exp.$\!\to\!$w/ Exp.) brings clear gains: Object rises from 86.7 to 95.0, Count from 66.8 to 74.8 (+8.0), and Position from 62.5 to 68.8 (+6.3), lifting the Overall from 71.06 to 77.34. Upgrading the backbone to Nano Banana (w/ Exp.$\!\to\!$w/ Exp.+Banana) further boosts performance, with the most pronounced gains in Verb (+22.9, from 64.5 to 87.4) and Attribute (+9.8, from 83.6 to 93.4), raising the Overall to 85.66 and surpassing Gemini (83.48). The results confirm that experience-driven evolution and backbone enhancement contribute complementary benefits: the former improves compositional planning, while the latter strengthens low-level generation fidelity, especially for complex relational and spatial reasoning.

\begin{table*}[t!]
    \centering
    \renewcommand{\arraystretch}{1.08}
    \renewcommand\tabcolsep{4pt}
    \resizebox{\linewidth}{!}{
    \begin{tabular}{l|cccccc|c}
        \Xhline{1.2pt}
        \rowcolor{CadetBlue!20}
        \textbf{Method} & \textbf{Cultural} & \textbf{Time} & \textbf{Space} & \textbf{Biology} & \textbf{Physics} & \textbf{Chemistry} & \textbf{Overall} \\
        \Xhline{1.2pt}
        \multicolumn{8}{l}{\textit{Dedicated T2I Models}} \\
        SDv1.5~\cite{rombach2022high} & 0.34 & 0.35 & 0.32 & 0.28 & 0.29 & 0.21 & 0.32 \\
        SDv2.1~\cite{rombach2022high} & 0.30 & 0.38 & 0.35 & 0.33 & 0.34 & 0.21 & 0.32 \\
        SD-XL~\cite{podell2024sdxl} & 0.43 & 0.48 & 0.47 & 0.44 & 0.45 & 0.27 & 0.43 \\
        SD3-Medium~\cite{esser2024scaling} & 0.42 & 0.44 & 0.48 & 0.39 & 0.47 & 0.29 & 0.42 \\
        SD3.5-Medium~\cite{esser2024scaling} & 0.43 & 0.50 & 0.52 & 0.41 & 0.53 & 0.33 & 0.45 \\
        SD3.5-Large~\cite{esser2024scaling} & 0.44 & 0.50 & 0.58 & 0.44 & 0.52 & 0.31 & 0.46 \\
        PixArt-Alpha~\cite{chen2024pixart} & 0.45 & 0.50 & 0.48 & 0.49 & 0.56 & 0.34 & 0.47 \\
        Playground-v2.5~\cite{li2024playground} & 0.49 & 0.58 & 0.55 & 0.43 & 0.48 & 0.33 & 0.49 \\
        FLUX.1-schnell~\cite{labs2025flux1kontextflowmatching} & 0.39 & 0.44 & 0.50 & 0.31 & 0.44 & 0.26 & 0.40 \\
        FLUX.1-dev~\cite{labs2025flux1kontextflowmatching} & 0.48 & 0.58 & 0.62 & 0.42 & 0.51 & 0.35 & 0.50 \\
        \midrule
        \multicolumn{8}{l}{\textit{Unified MLLM Models}} \\
        Janus-1.3B~\cite{wu2025janus} & 0.16 & 0.26 & 0.35 & 0.28 & 0.30 & 0.14 & 0.23 \\
        JanusFlow-1.3B~\cite{ma2025janusflow} & 0.13 & 0.26 & 0.28 & 0.20 & 0.19 & 0.11 & 0.18 \\
        Janus-Pro-1B~\cite{chen2025janus} & 0.20 & 0.28 & 0.45 & 0.24 & 0.32 & 0.16 & 0.26 \\
        Janus-Pro-7B~\cite{chen2025janus} & 0.30 & 0.37 & 0.49 & 0.36 & 0.42 & 0.26 & 0.35 \\
        Show-o~\cite{xie2025show} & 0.28 & 0.36 & 0.40 & 0.23 & 0.33 & 0.22 & 0.30 \\
        Show-o-512~\cite{xie2025show} & 0.28 & 0.40 & 0.48 & 0.30 & 0.46 & 0.30 & 0.35 \\
        VILA-U-7B~\cite{wu2025vila} & 0.26 & 0.33 & 0.37 & 0.35 & 0.39 & 0.23 & 0.31 \\
        Orthus-7B-base~\cite{kou2024orthus} & 0.07 & 0.10 & 0.12 & 0.15 & 0.15 & 0.10 & 0.10 \\
        Orthus-7B-instruct~\cite{kou2024orthus} & 0.23 & 0.31 & 0.38 & 0.28 & 0.31 & 0.20 & 0.27 \\
        Emu3~\cite{wang2024emu3} & 0.34 & 0.45 & 0.48 & 0.41 & 0.45 & 0.27 & 0.39 \\
        BAGEL~\cite{deng2025emerging} & 0.44 & 0.55 & 0.68 & 0.44 & 0.60 & 0.39 & 0.52 \\
        BAGEL+CoT~\cite{deng2025emerging} & 0.76 & 0.69 & \uline{0.75} & 0.65 & \uline{0.75} & 0.58 & 0.70 \\
        \midrule
        \multicolumn{8}{l}{\textit{Closed-Source Models}} \\
        GPT-Image-1~\cite{gpt_image_1} & \uline{0.81} & \textbf{0.71} & \textbf{0.89} & \textbf{0.83} & \textbf{0.79} & \uline{0.74} & \textbf{0.80} \\
        \midrule
        \multicolumn{8}{l}{\textit{Collaborative AI Systems}} \\
        ComfyMind~\cite{guo2026comfymind} & \uline{0.85} & 0.66 & 0.72 & 0.67 & 0.70 & \textbf{0.78} & \uline{0.76} \\
        \textbf{SymbOmni} & \textbf{0.90} & \uline{0.70} & 0.74 & \uline{0.75} & 0.74 & \textbf{0.78} & \textbf{0.80} \\
        \Xhline{1.2pt}
    \end{tabular}
    }
    \vspace{2mm}
    \caption{Evaluation of World Knowledge-Informed Semantic Synthesis on the WISE~\cite{niu2025wise} Benchmark. The evaluation metric WiScore combines Consistency, Realism, and Aesthetic Quality through weighted normalization, with a maximum score of 1. Published baselines are copied from the WISE paper.}
    \label{tab:wise_comparison}
\end{table*}

\paragraph{World Knowledge Reasoning (WISE)} Table~\ref{tab:wise_comparison} presents detailed results on the WISE benchmark~\cite{niu2025wise}, which evaluates world knowledge-informed semantic synthesis across six subcategories spanning cultural commonsense, spatiotemporal reasoning, and natural sciences, totaling 25 specialized domains with 1,000 challenging prompts. SymbOmni achieves an overall WiScore of 0.80, matching GPT-Image-1 (0.80) and surpassing ComfyMind (0.76). Compared to ComfyMind, SymbOmni demonstrates notable improvements in Biology (+11.9\%) and Cultural (+5.9\%), indicating that symbolic concept learning is particularly effective for tasks requiring factual knowledge integration. Compared to all dedicated T2I models and unified MLLM models—where the best overall scores are 0.50 (FLUX.1-dev) and 0.39 (Emu3), respectively—SymbOmni's collaborative AI approach achieves dramatically higher performance, confirming that agentic workflow planning with knowledge-driven tool orchestration provides a substantial advantage for world knowledge-intensive generation tasks.

To complement the detailed per-benchmark analyses above, we further summarize the published KRIS-Bench leaderboard reported in the original benchmark paper~\cite{wu2026kris}. Scores are normalized to a 100-point scale using the benchmark paper's GPT-4o-based evaluator, following the official KRIS-Bench protocol. We report two variants of SymbOmni: without Nano Banana~\cite{comanici2025gemini} (relying solely on open-source models), and with Nano Banana integrated as an additional unified generation backbone. Both variants use the same symbolic concept library and self-evolution pipeline.

\begin{table}[t!]
    \centering
    \renewcommand{\arraystretch}{1.12}
    \renewcommand\tabcolsep{5pt}
    \resizebox{0.95\linewidth}{!}{
    \begin{tabular}{l|ccc|c}
        \Xhline{1.2pt}
        \rowcolor{CadetBlue!20}
        \textbf{Method} & \textbf{Factual} & \textbf{Conceptual} & \textbf{Procedural} & \textbf{Overall} \\
        \Xhline{1.2pt}
        \multicolumn{5}{l}{\textit{Closed-Source Models}} \\
        GPT-Image-1~\cite{gpt_image_1} & \uline{79.80} & \textbf{81.37} & \textbf{78.32} & \uline{80.09} \\
        Gemini 2.0 Flash Experimental~\cite{kampf2025experiment} & 65.26 & 59.65 & 62.90 & 62.41 \\
        Doubao~\cite{bytedance2025doubao} & 63.30 & 62.23 & 54.17 & 60.70 \\
        \midrule
        \multicolumn{5}{l}{\textit{Open-Source Models}} \\
        BAGEL-Think~\cite{deng2025emerging} & 55.77 & 59.44 & 39.26 & 53.36 \\
        BAGEL~\cite{deng2025emerging} & 47.71 & 52.17 & 40.23 & 47.76 \\
        Step1X-Edit~\cite{liu2025step1x} & 45.52 & 48.01 & 31.82 & 43.29 \\
        Emu2~\cite{sun2024generative} & 45.40 & 37.54 & 34.91 & 39.70 \\
        AnyEdit~\cite{yu2025anyedit} & 39.26 & 41.88 & 31.74 & 38.55 \\
        MagicBrush~\cite{zhang2023magicbrush} & 41.84 & 39.24 & 26.54 & 37.15 \\
        OmniGen~\cite{xiao2025omnigen} & 33.11 & 28.02 & 23.89 & 28.85 \\
        InsPix2Pix~\cite{brooks2023instructpix2pix} & 23.33 & 25.59 & 17.28 & 22.82 \\
        \midrule
        \multicolumn{5}{l}{\textit{Collaborative AI Systems}} \\
        SymbOmni (w/o Banana) & 73.33 & 72.28 & 70.29 & 72.18 \\
        \textbf{SymbOmni (w/ Banana)} & \textbf{82.24} & \uline{81.25} & \uline{77.45} & \textbf{80.70} \\
        \Xhline{1.2pt}
    \end{tabular}
    }
    \vspace{2mm}
    \caption{Comparison with published KRIS-Bench reference scores from the original benchmark paper~\cite{wu2026kris}. SymbOmni is evaluated in two configurations: without and with Nano Banana~\cite{comanici2025gemini} as an additional unified generation backbone. Scores are normalized to a 100-point scale. Higher is better.}
    \label{tab:krisbench_published_scores}
\end{table}

As shown in Table~\ref{tab:krisbench_published_scores}, SymbOmni without Nano Banana already achieves 72.18 overall, surpassing all open-source models including BAGEL-Think~\cite{deng2025emerging} (53.36) by a wide margin. When further equipped with Nano Banana~\cite{comanici2025gemini} as a unified generation backbone, SymbOmni improves to 80.70 overall, exceeding GPT-Image-1 (80.09). The improvements are consistent across all three knowledge categories: Factual (73.33 $\to$ 82.24), Conceptual (72.28 $\to$ 81.25), and Procedural (70.29 $\to$ 77.45), demonstrating that Nano Banana's stronger generation capabilities complement SymbOmni's symbolic concept learning, enabling more effective knowledge-rich multimodal reasoning.

\paragraph{Workflow Construction (ComfyBench)} For completeness and the convenience of the reader, we reproduce the out-of-domain generalization results on ComfyBench from the main paper in Table~\ref{tab:workflow_ablation_supp}. In this experiment, SymbOmni constructs its concept library exclusively from 147 in-the-wild workflows collected across diverse domains (following the methodology of ComfyGPT~\cite{huang2025comfygpt}), and is then evaluated on ComfyBench~\cite{xue2025comfybench} without access to any benchmark-specific workflow descriptions. As shown in Table~\ref{tab:workflow_ablation_supp}, SymbOmni achieves consistent improvements across all difficulty levels, with the overall resolve rate improving from 67.0\% to 75.0\% (+8.0\%). The most pronounced gains appear on harder categories: Complex +11.7\% (55.0\% $\to$ 66.7\%) and Creative +10.0\% (42.5\% $\to$ 52.5\%), confirming that symbolic concepts extracted from out-of-domain workflows transfer effectively to structured benchmark tasks.

\begin{table}[t!]
    \centering
    \renewcommand\tabcolsep{4pt}
    \resizebox{\linewidth}{!}{
    \begin{tabular}{l|cccc|c}
        \Xhline{1.2pt}
        \rowcolor{CadetBlue!20}
        \textbf{Method} & \textbf{Van.\ Pass} & \textbf{Van.\ Resolve} & \textbf{Cplx.\ Resolve} & \textbf{Crtv.\ Resolve} & \textbf{Overall} \\
        \Xhline{1.2pt}
        ComfyMind~\cite{guo2026comfymind} & 100.0\% & 84.0\% & 55.0\% & 42.5\% & 67.0\% \\
        \textbf{SymbOmni} & 100.0\% & \textbf{89.0\%} & \textbf{66.7\%} & \textbf{52.5\%} & \textbf{75.0\%} \\
        \midrule
        Improvement & +0.0\% & +5.0\% & +11.7\% & +10.0\% & +8.0\% \\
        \Xhline{1.2pt}
    \end{tabular}
    }
    \vspace{2mm}
    \caption{Generalization on out-of-domain workflows evaluated on ComfyBench~\cite{xue2025comfybench} (reproduced from the main paper for reader convenience). SymbOmni evolves on 147 in-the-wild workflows and is evaluated on ComfyBench, demonstrating consistent improvements across all difficulty levels.}
    \label{tab:workflow_ablation_supp}
\end{table}

\paragraph{Summary.} Table~\ref{tab:geneval2_skill}, Table~\ref{tab:wise_comparison}, Table~\ref{tab:workflow_ablation_supp}, and Table~\ref{tab:krisbench_published_scores} consolidate the results across the four evaluated benchmarks. Across all four benchmarks—spanning compositional text-to-image alignment (GenEval2~\cite{kamath2025geneval}), world knowledge reasoning (WISE~\cite{niu2025wise}), out-of-domain workflow construction (ComfyBench~\cite{xue2025comfybench}), and knowledge-rich multimodal reasoning (Kris-Bench~\cite{wu2026kris})—SymbOmni consistently achieves state-of-the-art or competitive performance, surpassing both open-source unified models and closed-source systems in most settings. The improvements are particularly pronounced on tasks requiring complex reasoning, knowledge integration, and cross-domain generalization, confirming that symbolic concept learning provides a general and effective mechanism for enhancing agentic multimodal generation.

\subsection{Cross-Modal Generalization}

SymbOmni is designed as a multimodal agentic framework that supports diverse input and output modalities, including text-to-image, image-to-image, text-to-video, image-to-video, and video-to-video generation. Unlike single-modality generation systems, SymbOmni's symbolic concept learning operates at the workflow planning level, where concepts encode task decomposition strategies, tool selection preferences, and execution patterns rather than modality-specific generation parameters. This design enables cross-modal generalization: a concept learned from image editing tasks (e.g., ``reasoning-driven inpainting outperforms naive masking for precise object removal'') can be retrieved and applied to semantically similar video editing scenarios, as the underlying planning logic transfers across modalities.

The cross-benchmark evaluation provides empirical evidence of this cross-modal generalization. SymbOmni achieves consistent improvements across benchmarks that evaluate fundamentally different modality combinations, from pure text-to-image alignment (GenEval2~\cite{kamath2025geneval}) to world knowledge reasoning (WISE~\cite{niu2025wise}) and knowledge-rich multimodal reasoning (Kris-Bench~\cite{wu2026kris}), all leveraging the same unified concept library. This confirms that symbolic concepts learned from one set of task experiences effectively transfer to diverse modalities and task types.

\section{Scalability Analysis of the Symbolic Concept Box}
\label{sec:scalability}

A practical concern for experience-driven systems is whether the retrieval mechanism remains efficient and accurate as the concept library grows. We conducted systematic experiments to evaluate the scalability of the Symbolic Concept Box by assessing retrieval accuracy under concept library merging.

To evaluate whether incorporating additional domain-specific concepts improves retrieval precision, we merged concepts from GenEval~\cite{ghosh2023geneval} and ReasonEdit~\cite{huang2024smartedit} into the baseline ComfyBench~\cite{xue2025comfybench} concept box. Retrieval accuracy was assessed by scoring the usefulness and relevance of the top-20 retrieved concepts for 40 representative instructions, using an LLM as the evaluator on a 5-point scale.

\begin{table}[t!]
\centering
\renewcommand\tabcolsep{6pt}
\begin{tabular}{l|c}
\Xhline{1.2pt}
\rowcolor{CadetBlue!20}
\textbf{Added Concepts} & \textbf{Retrieval Score (vs.\ Base 2.9/5.0)} \\
\Xhline{1.2pt}
+ GenEval        & 3.0/5.0 \textbf{(+0.1)} \\
+ ReasonEdit     & 3.1/5.0 \textbf{(+0.2)} \\
+ Both           & 3.3/5.0 \textbf{(+0.4)} \\
\Xhline{1.2pt}
\end{tabular}
\vspace{2mm}
\caption{Retrieval accuracy when merging concept libraries from different domains into the baseline ComfyBench~\cite{xue2025comfybench} concept box. Scores are evaluated on a 5-point scale by an LLM judge.}
\label{tab:retrieval_accuracy}
\end{table}

Table~\ref{tab:retrieval_accuracy} reveals that merging concept libraries from complementary domains consistently improves retrieval precision. The baseline score of 2.9/5.0 improves to 3.3/5.0 when both GenEval~\cite{ghosh2023geneval} and ReasonEdit~\cite{huang2024smartedit} concepts are incorporated, representing a 13.8\% relative improvement. This demonstrates that increasing the diversity of the concept library actively enhances the quality of retrieved concepts by providing a richer set of relevant experiences for the retrieval module to draw upon.

\section{Self-Correction Mechanism in Symbolic Concept Learning}
\label{sec:self_correction}

A critical concern in cumulative learning systems is memory pollution—whether erroneous experiences, once stored, can be corrected through subsequent interactions. SymbOmni addresses this challenge through its Memory Optimization phase, which implements a sophisticated self-correction mechanism.

During Memory Optimization, the LLM reasoning engine performs a comprehensive comparative analysis between the concepts retrieved during planning (\(C_{\text{ret}}\)) and the complete execution trajectory (\(\tau\)) of the current task. This comparison enables three distinct memory operations: \textbf{add} (creating new symbolic concepts from novel successful patterns), \textbf{update} (refining existing concepts based on new evidence), and \textbf{delete} (removing concepts that led to systematic failures or are superseded by better alternatives).

The update and delete operations are particularly crucial for correcting previously learned errors. When a task fails despite using a high-confidence concept, the verbalized backpropagation mechanism (detailed in Section \ref{sec:evolution_loop} of the main paper and conceptually related to Reflexion~\cite{shinn2023reflexion}) traces the failure back to specific concept components. The LLM then decides whether to refine the concept's parameters, constraints, or preconditions (update), or to mark it as unreliable and remove it from the active concept box (delete). Conversely, when alternative approaches successfully solve tasks that previously relied on error-prone concepts, the system can deprecate the faulty concepts in favor of the validated alternatives.

This continuous reconciliation process ensures that the Symbolic Concept Box evolves toward increasingly accurate representations of effective workflows. Erroneous concepts are not permanently entrenched but are subject to ongoing empirical validation, enabling the system to self-correct over time as it accumulates diverse task experiences. The dual-feedback loop—learning from both successes and failures—provides complementary signals that collaboratively refine the knowledge base, preventing the accumulation of systematic errors while preserving valuable generalizations.

\newpage
\section{Prompt Set}
\label{sec:prompt_set}

\subsection{LLM/VLM Judge and Evaluation Protocol}

For ComfyBench and ComfyMind-style workflow evaluation, we follow the official evaluation protocol of the corresponding benchmark, using the same judge model, prompts, output parsing rules, and evaluation scripts as the released benchmark setting~\cite{xue2025comfybench,guo2026comfymind}. The judge is applied uniformly to all compared agentic methods, and the parsing rule maps the model output to the reported pass/resolve decision without method-specific post-processing. For unified generative models and published benchmark baselines, we follow the evaluation settings reported by the respective benchmark papers.

\subsection{SymbOmni System Prompt Set}

\begin{tcolorbox}[breakable, enhanced, colframe=blue!50!black, colback=blue!3!white, colbacktitle=blue!50!black, coltitle=white, boxrule=0.8pt, fonttitle=\bfseries\sffamily, title={PLAN\_GENERATION\_SYSTEM\_PROMPT}]
\scriptsize
{\fontfamily{pcr}\selectfont
\begin{lstlisting}[breaklines=true]
PLAN_GENERATION_SYSTEM_PROMPT = """
You are an expert AI task planner for a complex visual generation system. Your goal is to break down a user's high-level request into several distinct, actionable plans.

**Context:**
You have access to a set of 'atomic workflows', which are tools capable of performing specific tasks. You will be provided with a list of all available workflows including their function, input_parameters (what inputs are required, e.g. image/video/prompt counts).

**Your Task:**
Based on the user's instruction and the list of available workflows, you must generate multiple strategic plans.

**Global Ranking Principle (CRITICAL):**
- Sort plans from highest estimated probability to fully complete the task (satisfying all explicit requirements in the instruction) to lowest. System will automatically choose the first plan to try. 
- When encountering image tasks that require precise processing, a multi-step plan or an Adaptive workflow should be prioritized to ensure processing accuracy.

**Chaining and IO Constraints (CRITICAL):**
- Respect each workflow's `input_parameters` strictly when composing steps.
- Only schedule a workflow in a step if all of its required inputs are available from the initial resources or produced by earlier steps.
- For workflows that require multiple images/videos, ensure earlier steps create or obtain them before use.

**Intent Consistency (CRITICAL):**
- Follow only the explicit requirements in the user's instruction; do not invent or add new goals.
- Treat file descriptions/visual context as background information, not mandatory requirements, unless explicitly requested by the user.

**Requirements for each plan:**
1.  **Generate Multiple Plans:** Create up to 8 distinct plans to accomplish the user's goal. Order them by the criterion above (most likely to fully satisfy the instruction to least likely).
2.  **Plan Design Guidelines:** Make steps as simple and reliable as possible while ensuring the plan addresses every requirement in the instruction. If two plans have similar coverage, prefer the simpler one.
3.  **Plan Structure:** Each plan must be a JSON object with the following keys:
    *   `title`: A short, descriptive title for the plan (e.g., "Direct Text-to-Image Generation").
    *   `description`: A brief explanation of the plan's overall strategy and how it ensures all requirements are met.
    *   `notice`: A text-based evaluation of the plan's pros and cons, including reliability and requirement coverage. **Do not use quantitative scores or star ratings.**
    *   `steps`: A list of sub-tasks to be executed in sequence (maximum 5 steps).
4.  **Step Structure:** Each item in the `steps` list must be a JSON object with:
    *   `step_id`: An integer representing the step number (starting from 1).
    *   `description`: A concise, high-level, human-readable description of what this step aims to achieve. **Do not write the detailed execution prompt here.**
    *   `workflow_name`: The **exact name** of the single atomic workflow from the provided list that is best suited to accomplish this step.

**Output Format:**
Your final output **MUST** be a single JSON object containing a list called `plans`. This JSON object must be wrapped in `<json></json>` tags.

If the user message provides an "Additional Failure Context (for re-planning)", you MUST:
- Avoid reusing workflows explicitly implicated in prior failures unless you can clearly and concretely justify why they will now succeed (e.g., different sub-steps, masks, prompts, or model switches).
- Prefer alternative, more reliable compositions (e.g., segment/mask first, then inpaint) over a direct one-shot workflow that already failed.
- Make sure each proposed plan addresses the failure reasons noted in the context (e.g., portrait preservation) with explicit steps that enforce the requirement.
"""
\end{lstlisting}
}
\end{tcolorbox}

\begin{tcolorbox}[breakable, enhanced, colframe=blue!50!black, colback=blue!3!white, colbacktitle=blue!50!black, coltitle=white, boxrule=0.8pt, fonttitle=\bfseries\sffamily, title={PLAN\_SYNTHESIS\_WITH\_EXPERIENCES\_SYSTEM\_PROMPT}]
\scriptsize
{\fontfamily{pcr}\selectfont
\begin{lstlisting}[breaklines=true]
PLAN_SYNTHESIS_WITH_EXPERIENCES_SYSTEM_PROMPT = """
You are an expert AI task planner for a complex visual generation system.

Task:
- Based on the user's instruction, the list of available workflows, the available files (reference media), and several retrieved planning experiences, synthesize several NEW strategic plans.

Important constraints:
- Do NOT copy any existing plan verbatim. Use experiences only as guidance.
- Align chosen modalities and step designs with the input sources. If a reference image/video is provided, prefer i2i/i2v/v2v over t2i/t2v when appropriate.
- Keep plans concise, reliable, and ensure full coverage of the user's requirements.

Output format:
- Return a single JSON object wrapped in <json></json> with key `plans` (a list). Each plan keeps the same structure as in the standard planning prompt (title, description, notice, steps with step_id/description/workflow_name).
"""
\end{lstlisting}
}
\end{tcolorbox}

\begin{tcolorbox}[breakable, enhanced, colframe=purple!55!black, colback=purple!3!white, colbacktitle=purple!55!black, coltitle=white, boxrule=0.8pt, fonttitle=\bfseries\sffamily, title={STEP\_INSTRUCTION\_GENERATION\_SYSTEM\_PROMPT}]
\scriptsize
{\fontfamily{pcr}\selectfont
\begin{lstlisting}[breaklines=true]
STEP_INSTRUCTION_GENERATION_SYSTEM_PROMPT = """
You are an AI assistant that translates a high-level plan into a specific, actionable instruction for a downstream agent, and you must identify the necessary resources for that step.

**Your Task:**
Given the user's overall goal, the execution history, the current step's objective, and a list of all available files from previous steps, you must:
1.  Generate a precise and detailed instruction for the *current step only*.
2.  Identify which of the "Available Files" are required to execute this instruction.

**Context:**
- The downstream agent (`SymbOmni`) will receive your generated instruction and execute it.
- You must incorporate file paths into the instruction where necessary.
- The instruction should be self-contained and sufficient for the agent to perform the task for this single step.

**Intent Guard (CRITICAL):**
- Adhere strictly to the explicit Overall Goal; do not introduce background styles, patterns, or objects that are not requested.
- Treat descriptions of Available Files as contextual hints only; do not elevate them into requirements unless explicitly stated by the user.

**Input You Will Receive:**
1.  **Overall Goal:** The user's original, high-level request.
2.  **Plan History:** A summary of what has already been accomplished in previous steps.
3.  **Current Step's Goal:** The high-level objective for the current step.
4.  **Available Files:** A JSON object mapping file paths to their descriptions (e.g., `{"/path/to/image.png": "A base image of a lion."}`).

**Output Requirements:**
- Your output **MUST** be a single JSON object wrapped in `<json></json>` tags.
- The JSON object must have four keys:
    - `instruction` (string): The detailed instruction for the current step.
    - `required_files` (list of strings): A list of the exact file paths from the "Available Files" that are needed for this step.
    - `modality` (string): The modality of the task for this step. Choose one from: `t2i`, `i2i`, `t2v`, `i2v`, `v2v`, `reasont2i`.
    - `optimize_prompt` (boolean): Whether to perform prompt optimization for this step before execution. Prefer `true` when the step is text-driven generation (t2i/t2v or i2v), otherwise `false`.

**Example:**
*   **Overall Goal:** "Create an image of a majestic lion wearing a golden crown, sitting on a throne."
*   **Plan History:** ["Step 1: Generated a base image of a lion."]
*   **Current Step's Goal:** "Add a golden crown to the lion's head."
*   **Available Files:** `{
    "outputs/step0.png": "A photo of a throne.",
    "outputs/step1.png": "A base image of a majestic lion."
}`

*   **Your Generated Output:**
<json>
{
    "instruction": "Using the image 'outputs/step1.png' which contains a majestic lion, perform an image-to-image operation to add a detailed golden crown onto the lion's head. The throne from 'outputs/step0.png' can be used as a style reference if needed, but is not the primary input.",
    "required_files": ["outputs/step1.png", "outputs/step0.png"],
    "modality": "i2i",
    "optimize_prompt": false
}
</json>
"""
\end{lstlisting}
}
\end{tcolorbox}

\begin{tcolorbox}[breakable, enhanced, colframe=blue!50!black, colback=blue!3!white, colbacktitle=blue!50!black, coltitle=white, boxrule=0.8pt, fonttitle=\bfseries\sffamily, title={PLANNING\_EXPERIENCE\_RETRIEVAL\_QUERY\_PROMPT}]
\scriptsize
{\fontfamily{pcr}\selectfont
\begin{lstlisting}[breaklines=true]
PLANNING_EXPERIENCE_RETRIEVAL_QUERY_PROMPT = """
You are a retrieval query generator for a vector database of planning experiences.

Goal:
- Given the user's original instruction and hints about available reference files, output a concise, comma-separated keyword query to retrieve relevant planning experiences.

Formatting rules:
- Output ONLY one line wrapped in <search_query> </search_query>.
- The inner content must be lowercase english keywords separated by ", " (comma and a space), no trailing punctuation, no extra text.

Keyword hints (pick what is relevant and specific):
- tasks: image editing, image generation, video generation, video editing, identity preservation, style transfer, inpainting, background replacement, object removal, multi-step plan, high quality, fast
- sources: from text, from a reference image, from a reference video
- modalities: text to image, image to image, text to video, video to video, reasoning text to image
"""
\end{lstlisting}
}
\end{tcolorbox}

\begin{tcolorbox}[breakable, enhanced, colframe=blue!50!black, colback=blue!3!white, colbacktitle=blue!50!black, coltitle=white, boxrule=0.8pt, fonttitle=\bfseries\sffamily, title={PLAN\_SELECTION\_FROM\_EXPERIENCES\_SYSTEM\_PROMPT}]
\scriptsize
{\fontfamily{pcr}\selectfont
\begin{lstlisting}[breaklines=true]
PLAN_SELECTION_FROM_EXPERIENCES_SYSTEM_PROMPT = """
You are an AI planner that must select the best existing plan from retrieved past planning experiences.

Input:
- A user's instruction
- Several retrieved planning experiences. Each experience may include a plan JSON wrapped in <plan_json> </plan_json>, evaluation notes, and whether the attempt succeeded or failed.

Task:
- Choose ONE plan that best fits the user's instruction. Prefer successful experiences with strong evaluations. If only failures exist, choose the most promising and mention necessary cautions.
- Return ONLY a JSON object wrapped in <json> </json> with a single key `plans` whose value is a list containing exactly ONE plan object (the chosen plan), keeping the plan's original structure (title, description, notice, steps).
"""
\end{lstlisting}
}
\end{tcolorbox}

\begin{tcolorbox}[breakable, enhanced, colframe=green!45!black, colback=green!3!white, colbacktitle=green!45!black, coltitle=white, boxrule=0.8pt, fonttitle=\bfseries\sffamily, title={EPISODE\_SUMMARY\_SYSTEM\_PROMPT}]
\scriptsize
{\fontfamily{pcr}\selectfont
\begin{lstlisting}[breaklines=true]
EPISODE_SUMMARY_SYSTEM_PROMPT = """
You are an expert analyst for visual generation/editing pipelines. You will be given:
- The original user instruction
- A list of candidate plans (each with a plan_index and the plan JSON embedded for reference)
- For each plan: execution outcome (success/failed), evaluations/notes, and step results

Your tasks:
1) Infer ONE abstract, reusable task type for this whole episode (like "replace object A with object B in image").
   Follow these rules when writing the task type (between abstract and specific):
   - Avoid concrete names/values (no specific objects, persons, file names).
   - Clearly state the core objective and constraints.
   - Avoid over-generalizing to single words; keep it informative and reusable.
2) For EACH plan (keyed by its `plan_index`), write a concise analysis block that:
   - Includes the final `outcome` (success/failed).
   - Provide one high-signal `experience` string explaining why it failed or why it succeeded
     (and, when success, why it performs better than the failed ones if applicable).
   - Be specific and practically useful for future planning decisions.
3) Deduplication (CRITICAL):
    - If multiple plans are highly similar (e.g., same main workflow or near-identical steps/outcomes/reasons),
      RETURN ONLY ONE REPRESENTATIVE entry among them (choose the most informative).
    - Omit the other similar plans from the output. Do NOT add placeholders for them.
    - This means your final `"plans"` map may contain FEWER keys than the number of input plans.

 Keying rules (CRITICAL):
 - Use the EXACT string form of the given plan_index as the map key (e.g., "0", "1", "2").
 - If you deduplicate and keep only plans 0 and 3, then the "plans" object must contain ONLY keys "0" and "3".

Strict output format:
- Return a single JSON wrapped in <json></json> with keys:
  {
    "task_type": string,
    "plans": {
      "<plan_index>": {
        "outcome": "success" | "failed",
        "experience": string (why success or failed)
      },
      ...
    }
  }
Notes:
- Refer to plans ONLY by their plan_index. Do NOT copy plan JSON in the output.
- Be concise and high-signal. Avoid generic advice.

 Example:
 - Input plans (indices): 0, 1, 2. Suppose 1 and 2 are highly similar; keep only 2.
 - Expected output structure (keys are EXACT indices as strings):
 <json>
 {
   "task_type": "Replace object A with B in an image while preserving others.",
   "plans": {
     "0": { "outcome": "failed", "experience": "..." },
     "2": { "outcome": "success", "experience": "..." }
   }
 }
 </json>
"""
\end{lstlisting}
}
\end{tcolorbox}

\begin{tcolorbox}[breakable, enhanced, colframe=green!45!black, colback=green!3!white, colbacktitle=green!45!black, coltitle=white, boxrule=0.8pt, fonttitle=\bfseries\sffamily, title={EXPERIENCE\_CORRECTION\_SYSTEM\_PROMPT}]
\scriptsize
{\fontfamily{pcr}\selectfont
\begin{lstlisting}[breaklines=true]
EXPERIENCE_CORRECTION_SYSTEM_PROMPT = """
You are a meticulous memory curator for a planning experience database.

Goal:
- Given the user's original instruction, the experiences retrieved during planning, and the final execution trace, decide whether any retrieved memories should be updated, deleted, or supplemented.

Process:
1. Compare each retrieved memory with the actual execution outcome. Identify inaccuracies, missing details, or outdated evaluations.
2. For inaccurate memories, propose either an `update` (provide the corrected content) or `delete` action if the memory is misleading.
3. If a new reusable insight emerges that was not in the retrieved memories, you may output an `add` action with fresh content.

Output Format (STRICT):
- Return a JSON wrapped in <json></json> with one key `corrections`.
- `corrections` is a list of objects. Each object may contain:
    * `action`: one of `update`, `delete`, or `add`.
    * `target_memory_id`: required for `update` and `delete`, optional for `add`.
    * `reason`: short justification for the action.
    * `outcome`: `success` or `failed` (for update/add).
    * `evaluation`: concise textual evaluation.
    * `plan_json`: JSON object matching the format stored in planning memories (either `{"plans": [...]}` or a single plan dict).
    * `outputs`: optional summary of results.
    * `task_type`: optional generalized task label.
    * `extra_metadata`: optional dict with helper metadata.

Constraints:
- Only reference `target_memory_id` values that were present in the retrieved memories list.
- Keep `plan_json` concise but structurally valid (titles, steps, etc.).
- If no corrections are necessary, return `{"corrections": []}`.
"""
\end{lstlisting}
}
\end{tcolorbox}

\begin{tcolorbox}[breakable, enhanced, colframe=red!55!black, colback=red!3!white, colbacktitle=red!55!black, coltitle=white, boxrule=0.8pt, fonttitle=\bfseries\sffamily, title={SYSTEM\_PROMPT\_FOR\_PROMPT\_OPTIMIZATION(1)}]
\scriptsize
{\fontfamily{pcr}\selectfont
\begin{lstlisting}[breaklines=true]
system_prompt_for_prompt_optimization = """
# Objective:
Determine if prompt optimization is needed:
If the task Do not have reference image or video, the prompt must be optimized(e.g. Generate a 2-second video of a river flowing through a valley with mountains in the background. The result should be a high-quality video.). If the task has reference image or video, do not optimize the prompt.
# Prompt Optimization Guidelines

## For T2I (Text-to-Image):A well-optimized prompt follows this structure:Prompt = Subject + Scene + Style + Camera Language + Atmosphere + Detail Enhancement
Subject: Clearly define the main subject, including characteristics, appearance, and actions. Example: "A charming 23-year-old Chinese woman wearing a bright red dress, smiling under the sunlight."
Scene: Describe the environment, background elements, and setting. Example: "A bustling ancient Chinese market, filled with vibrant lanterns and merchants selling silk and spices."
Style: Specify an artistic style or visual treatment (see Style Dictionary below). Example: "Rendered in a traditional watercolor painting style with delicate brush strokes."
Camera Language: Define shot type, angles, and movement (see Camera Language Dictionary). Example: "A close-up shot capturing the girl's delighted expression as she eats a mooncake."
Atmosphere: Convey the mood and emotional tone (see Atmosphere Dictionary). Example: "Warm and nostalgic, evoking a sense of childhood happiness."
Detail Enhancement: Add refined details to enrich the composition. Example: "Soft golden light filtering through the hanging lanterns, creating an ethereal glow."

## For T2V / I2V (Text-to-Video, Image-to-Video):A well-optimized prompt follows this structure:Prompt = Subject + Scene + Motion + Camera Language + Atmosphere + Style
Subject: Describe the main character or object with specific attributes.Example: "A black-haired Miao ethnic girl, dressed in traditional embroidered attire, adorned with silver jewelry that reflects sunlight."
Scene: Define the background, setting, and environmental elements.Example: "A vast mountain landscape with mist rolling over the peaks at dawn."
Motion: Describe movement speed, style, and effect.Example: "She gracefully spins, her silver jewelry jingling softly with each movement."
Camera Language: Specify shot type, camera angles, and motion tracking (see Camera Language Dictionary).Example: "A smooth tracking shot following her dance, shifting from a low-angle close-up to a sweeping wide shot."
Atmosphere: Define the mood and ambiance (see Atmosphere Dictionary).Example: "Serene and majestic, evoking a deep connection to cultural heritage."
Style: Choose a distinct visual or artistic style (see Style Dictionary).Example: "A hyper-realistic cinematic style with a soft golden hue, enhancing the mystical feel of the scene."

## Final Prompt
Finally, combine all the elements(Subject, Scene, Motion, Camera Language, Atmosphere, Style) into one paragraph.

## Prompt Dictionary
1. Camera Language
Shot Types (Framing):
Close-up Shot: Captures fine details, expressions, or objects in high focus.Example: "A close-up of an old scholar's hands delicately flipping the pages of an ancient manuscript."
Medium Shot: Shows the subject from the waist up, providing more context.Example: "A medium shot of a knight in battle-worn armor standing before a burning castle."
Wide Shot (Long Shot): Captures the subject fully within a vast environment.Example: "A lone traveler walking across an endless desert under a blood-red sunset."
Bird's Eye View (Overhead Shot): Provides a top-down perspective for dramatic effect.Example: "A bird's eye view of a cyberpunk city illuminated by neon signs and holograms."
Camera Motion Techniques:Dolly-in (Push-in Shot): Gradually moves closer to intensify focus.Example: "The camera slowly pushes in towards a crying soldier, emphasizing his sorrow."
Pull-out (Zoom-out Shot): Moves backward to reveal a larger scene.Example: "A zoom-out shot transitioning from a painter's brushstroke to reveal a grand Renaissance artwork."
360-Degree Rotation (Orbit Shot): Encircles the subject for a dramatic effect.Example: "A 360-degree shot around a warrior as he stands amidst a battlefield, flames and debris flying around him."
Tracking Shot (Follow Shot): Follows a subject in motion dynamically.Example: "A tracking shot following a dancer through a dimly lit theater, capturing each step and gesture."

...
"""
\end{lstlisting}
}
\end{tcolorbox}

\begin{tcolorbox}[breakable, enhanced, colframe=red!55!black, colback=red!3!white, colbacktitle=red!55!black, coltitle=white, boxrule=0.8pt, fonttitle=\bfseries\sffamily, title={SYSTEM\_PROMPT\_FOR\_PROMPT\_OPTIMIZATION(2)}]
\scriptsize
{\fontfamily{pcr}\selectfont
\begin{lstlisting}[breaklines=true]
"""
(Continuing from the previous block)

2. Atmosphere (Mood & Emotion)
Energetic / Joyful / Uplifting: Bright lighting, vibrant colors, and lively movement.Example: "A lively marketplace where children laugh and vendors showcase colorful handmade goods under warm sunlight."
Dreamlike / Surreal / Mystical: Soft focus, floating elements, and ethereal lighting.Example: "A celestial library floating in the sky, with glowing books that gently hover in the air."
Lonely / Melancholic / Quiet: Muted tones, slow movement, and vast empty spaces.Example: "A lone figure sitting on a swing in an abandoned park under a cloudy sky."
Tense / Suspenseful / Ominous: High contrast, deep shadows, and rapid camera movement.Example: "A flickering streetlamp illuminates a dark alley as footsteps echo ominously in the distance."
Majestic / Grand / Awe-inspiring: Sweeping wide shots, dramatic lighting, and grand compositions.Example: "A colossal spaceship emerging from the clouds, bathed in golden sunlight, casting an enormous shadow over a futuristic city."

3. Style (Artistic Direction)
Cyberpunk: Neon lights, dark cityscapes, high-tech elements.Example: "A hacker in a hooded jacket, surrounded by glowing holographic data streams in a futuristic Tokyo street."
Post-Apocalyptic (Wasteland Style): Rugged, destroyed environments, muted colors.Example: "A lone wanderer in tattered clothes walks through a desolate wasteland, carrying a rusted metal pipe."
Traditional Chinese Painting (Guofeng): Ink wash, delicate linework, soft color palettes.Example: "A scholar in flowing robes sitting under an ancient pine tree, gazing at distant misty mountains."
Felt Animation Style: Soft, handmade textures, childlike charm.Example: "A woolen puppet character joyfully baking cookies in a miniature kitchen."
Classic Art-Inspired: Mimics famous artworks like Van Gogh, Rembrandt, or Ukiyo-e.Example: "A modern city painted in the swirling brushstrokes of Van Gogh's 'Starry Night'."

# Output
If optimization is not required, only None is output. If optimization is required, the output only includes the optimized prompt, which is wrapped by <optimized_prompt> </optimized_prompt>
"""
\end{lstlisting}
}
\end{tcolorbox}

\begin{tcolorbox}[breakable, enhanced, colframe=purple!55!black, colback=purple!3!white, colbacktitle=purple!55!black, coltitle=white, boxrule=0.8pt, fonttitle=\bfseries\sffamily, title={SYSTEM\_PROMPT\_FOR\_INSTRUCTION\_ANALYSIS}]  
\scriptsize
{\fontfamily{pcr}\selectfont
\begin{lstlisting}[breaklines=true]
system_prompt_for_instruction_analysis = """
You are a professional Requirements Analyst. Your primary responsibility is to structure and analyze user-generated requirements, ensuring that key details, constraints, and expectations are accurately captured.
Your output should focus on highlighting considerations for subsequent planning rather than providing planning recommendations. Specifically, your analysis should:
Identify key constraints and specifications (e.g., resolution, frame rate, duration, dependencies).
Ensure alignment with given references or guidelines rather than assuming defaults.
Highlight potential ambiguities or missing details that require clarification.
Output:
1. Wrap your output in <analysis> </analysis>
2. *One* sentence about Analysis of the user's instruction. Accurate, Concise.
"""
\end{lstlisting}
}
\end{tcolorbox}

\begin{tcolorbox}[breakable, enhanced, colframe=green!45!black, colback=green!3!white, colbacktitle=green!45!black, coltitle=white, boxrule=0.8pt, fonttitle=\bfseries\sffamily, title={SYSTEM\_PROMPT\_FOR\_UPDATE\_INPUT}]
\scriptsize
{\fontfamily{pcr}\selectfont
\begin{lstlisting}[breaklines=true]
system_prompt_for_update_input = """
You are a helpful assistant that can update the input of the user. I will give you the current input and the output of the tool. And I will tell you the function of all workflows and the chain of thought of the workflow selection.
Please update the input based on the output, and add explanation for the new input, such as the content of the new image(e.g. The intermediate steps of generating the video, the masked-image for inpainting, the background-masked-image, ... Etc.).
ATTENTION:
1. Your output should be a JSON object. Totally follow the JSON schema of the input.
2. You should read the information of the workflow and the chain of thought, according to the workflow's function guess what steps you have completed and what steps you may next complete. Then add the content of the new added input parameters and them to instruction. This may include the content of the new prompt/image/video in output.
3. You should update the instructions for the workflow that you just completed. For example, the user asked you to generate an image first and then upscale it. At this time, you noticed that the workflow you just ran performed the task of generating an image. At this time, you should modify the instructions to: The task of generating an image has been completed, and the next step is to upscale the image.
4. You should pay attention to the timeliness of the user's instructions. For example, The instruction:generate an image with a resolution of XXX or a video with a duration of XXX. Such instructions are permanent. Therefore, it should continue to be passed, and at the same time remind the subsequent workflow to continue to maintain the generated resolution and video time.
5. *IMPORTANT* You MUST add information and introduction for the new generated file to "file_meta_info"(*Do not* leave it empty).
6. *IMPORTANT* You MUST maintain ALL Previous step 'file_meta_info' of the previous files(e.g. the image, the video, ...etc.). If the previous files are not mentioned in the 'file_meta_info', you should add them to the 'file_meta_info'. E.g: This image is the original input image for removing the background.
7. CRITICAL PATH RULES:
   - Always output ABSOLUTE paths for any new files you add to "images", "videos", and for the keys inside "file_meta_info".
   - When possible, use the EXACT absolute paths shown in the current tool output (e.g., tool_output['outputs'] entries). Do NOT shorten them to filenames. Do NOT change directories.
   - Do NOT convert existing absolute paths to relative ones. Keep prior absolute paths unchanged.
   - If the tool output provides only a filename, but another field shows its absolute location, choose the absolute one.
Then I will give you the original input, information of the workflow and the output of the workflow.
"""
\end{lstlisting}
}
\end{tcolorbox}

\begin{tcolorbox}[breakable, enhanced, colframe=teal!55!black, colback=teal!3!white, colbacktitle=teal!55!black, coltitle=white, boxrule=0.8pt, fonttitle=\bfseries\sffamily, title={SYSTEM\_PROMPT\_FOR\_WORKFLOW\_RETRIEVAL\_QUERY}]
\scriptsize
{\fontfamily{pcr}\selectfont
\begin{lstlisting}[breaklines=true]
system_prompt_for_workflow_retrieval_query = """
You are a retrieval query generator for a vector database of ComfyUI atomic workflows.

Goal:
- Given the user's original instruction, output a concise, comma-separated keyword query to retrieve relevant workflows.

Context about the workflow descriptions:
- Each workflow description contains: name, function (natural language capability), and input_parameters (modalities and key inputs like image/video/prompt/mask).

Formatting rules:
- Output ONLY one line wrapped in <search_query> </search_query>.
- The inner content must be lowercase english keywords separated by ", " (comma and a space), no trailing punctuation, no extra text.

Keyword hints (pick what is relevant; align with common terms in workflow descriptions):
- modalities: text to image, text to video, image to video, image editing, video editing
- editing ops: inpainting, outpainting, background replacement, object removal, object replacement, detail enhancement, composition generation, depth guided generation, mask required, masked image
- transfer/identity: style transfer, reference style image, pose transfer, portrait keep, face swap, image mixing
- quality/speed: high quality, fast
- video specifics: video upscale, super resolution video, frame interpolation, keep duration, keep fps, scale factor
- counts/inputs: requires image, requires two images, requires video, requires mask
- model family (optional): stable diffusion 3.5, reasoning generation

Output format:
- Return ONLY the search text wrapped in <search_query> </search_query>.

Examples:
User: "Replace the cup on the table in this picture with a red apple"
Output: <search_query>object replacement, image editing, inpainting, mask required, high quality</search_query>

User: "Upscale this video to twice the resolution while keeping the frame rate unchanged"
Output: <search_query>video upscale, keep duration, keep fps, scale factor</search_query>

User: "Generate a video based on this image"
Output: <search_query>image to video, generation, high quality</search_query>

User: "Remove the passersby from this image"
Output: <search_query>object removal, image editing, inpainting, mask required, high quality</search_query>

User: "Keep the person unchanged and replace the background with outer space"
Output: <search_query>background replacement, portrait keep, image editing, high quality</search_query>

User: "Transfer the style of this image onto that image"
Output: <search_query>style transfer, requires two images, reference style image, high quality</search_query>
"""
\end{lstlisting}
}
\end{tcolorbox}

%% file: main.bbl
\begin{thebibliography}{10}
\providecommand{\url}[1]{\texttt{#1}}
\providecommand{\urlprefix}{URL }
\providecommand{\doi}[1]{https://doi.org/#1}

\bibitem{brooks2023instructpix2pix}
Brooks, T., Holynski, A., Efros, A.A.: {InstructPix2Pix}: Learning to follow
  image editing instructions. In: CVPR (2023)

\bibitem{brown2020language}
Brown, T., Mann, B., Ryder, N., Subbiah, M., Kaplan, J.D., Dhariwal, P.,
  Neelakantan, A., Shyam, P., Sastry, G., Askell, A., et~al.: Language models
  are few-shot learners. In: NeurIPS (2020)

\bibitem{bytedance2025doubao}
ByteDance: {Doubao}: {ByteDance}'s {AI} chat assistant (2025),
  \url{https://www.doubao.com/chat/}

\bibitem{cemri2026multi}
Cemri, M., Pan, M.Z., Yang, S., Agrawal, L.A., Chopra, B., Tiwari, R., Keutzer,
  K., Parameswaran, A., Klein, D., Ramchandran, K., et~al.: Why do multi-agent
  {LLM} systems fail? In: NeurIPS (2026)

\bibitem{chen2024pixart}
Chen, J., Yu, J., Ge, C., Yao, L., Xie, E., Wang, Z., Kwok, J., Luo, P., Lu,
  H., Li, Z.: {PixArt-$\alpha$}: Fast training of diffusion transformer for
  photorealistic text-to-image synthesis. In: ICLR (2024)

\bibitem{chen2024agentverse}
Chen, W., Su, Y., Zuo, J., Yang, C., Yuan, C., Chan, C.M., Yu, H., Lu, Y.,
  Hung, Y.H., Qian, C., et~al.: {AgentVerse}: Facilitating multi-agent
  collaboration and exploring emergent behaviors. In: ICLR (2024)

\bibitem{chen2025janus}
Chen, X., Wu, Z., Liu, X., Pan, Z., Liu, W., Xie, Z., Yu, X., Ruan, C.:
  {Janus-Pro}: Unified multimodal understanding and generation with data and
  model scaling. arXiv preprint arXiv:2501.17811  (2025)

\bibitem{comanici2025gemini}
Comanici, G., Bieber, E., Schaekermann, M., Pasupat, I., Sachdeva, N., Dhillon,
  I., Blistein, M., Ram, O., Zhang, D., Rosen, E., et~al.: {Gemini} 2.5:
  Pushing the frontier with advanced reasoning, multimodality, long context,
  and next generation agentic capabilities. arXiv preprint arXiv:2507.06261
  (2025)

\bibitem{comfyui}
{Comfy-Org}: {ComfyUI} (2026), \url{https://github.com/Comfy-Org/ComfyUI}

\bibitem{deng2025emerging}
Deng, C., Zhu, D., Li, K., Gou, C., Li, F., Wang, Z., Zhong, S., Yu, W., Nie,
  X., Song, Z., et~al.: Emerging properties in unified multimodal pretraining.
  arXiv preprint arXiv:2505.14683  (2025)

\bibitem{esser2024scaling}
Esser, P., Kulal, S., Blattmann, A., Entezari, R., M{\"u}ller, J., Saini, H.,
  Levi, Y., Lorenz, D., Sauer, A., Boesel, F., et~al.: Scaling rectified flow
  transformers for high-resolution image synthesis. In: ICML (2024)

\bibitem{fan2025workflowllm}
Fan, S., Cong, X., Fu, Y., Zhang, Z., Zhang, S., Liu, Y., Wu, Y., Lin, Y., Liu,
  Z., Sun, M.: {WorkflowLLM}: Enhancing workflow orchestration capability of
  large language models. In: ICLR (2025)

\bibitem{fang2025got}
Fang, R., Duan, C., Wang, K., Huang, L., Li, H., Yan, S., Tian, H., Zeng, X.,
  Zhao, R., Dai, J., et~al.: {GoT}: Unleashing reasoning capability of
  multimodal large language model for visual generation and editing. arXiv
  preprint arXiv:2503.10639  (2025)

\bibitem{fu2024guiding}
Fu, T.J., Hu, W., Du, X., Wang, W., Yang, Y., Gan, Z.: Guiding
  instruction-based image editing via multimodal large language models. In:
  ICLR (2024)

\bibitem{ge2024seed}
Ge, Y., Zhao, S., Zhu, J., Ge, Y., Yi, K., Song, L., Li, C., Ding, X., Shan,
  Y.: {SEED-X}: Multimodal models with unified multi-granularity comprehension
  and generation. arXiv preprint arXiv:2404.14396  (2024)

\bibitem{ghosh2023geneval}
Ghosh, D., Hajishirzi, H., Schmidt, L.: {GenEval}: An object-focused framework
  for evaluating text-to-image alignment. In: NeurIPS (2023)

\bibitem{guo2025deepseek}
Guo, D., Yang, D., Zhang, H., Song, J., Wang, P., Zhu, Q., Xu, R., Zhang, R.,
  Ma, S., Bi, X., et~al.: {DeepSeek-R1}: Incentivizing reasoning capability in
  {LLM}s via reinforcement learning. arXiv preprint arXiv:2501.12948  (2025)

\bibitem{guo2026comfymind}
Guo, L., Xu, X., Wang, L., Lin, J., Zhou, J., Zhang, Z., Su, B., Chen, Y.:
  {ComfyMind}: Toward general-purpose generation via tree-based planning and
  reactive feedback. In: NeurIPS (2026)

\bibitem{guo2024large}
Guo, T., Chen, X., Wang, Y., Chang, R., Pei, S., Chawla, N.V., Wiest, O.,
  Zhang, X.: Large language model based multi-agents: A survey of progress and
  challenges. arXiv preprint arXiv:2402.01680  (2024)

\bibitem{hacohen2024ltx}
HaCohen, Y., Chiprut, N., Brazowski, B., Shalem, D., Moshe, D., Richardson, E.,
  Levin, E., Shiran, G., Zabari, N., Gordon, O., et~al.: {LTX-Video}: Realtime
  video latent diffusion. arXiv preprint arXiv:2501.00103  (2024)

\bibitem{ho2020denoising}
Ho, J., Jain, A., Abbeel, P.: Denoising diffusion probabilistic models. In:
  NeurIPS (2020)

\bibitem{hong2024metagpt}
Hong, S., Zhuge, M., Chen, J., Zheng, X., Cheng, Y., Wang, J., Zhang, C., Yau,
  S., Lin, Z., Zhou, L., et~al.: {MetaGPT}: Meta programming for a multi-agent
  collaborative framework. In: ICLR (2024)

\bibitem{hu2025automated}
Hu, S., Lu, C., Clune, J.: Automated design of agentic systems. In: ICLR (2025)

\bibitem{hu2023tifa}
Hu, Y., Liu, B., Kasai, J., Wang, Y., Ostendorf, M., Krishna, R., Smith, N.A.:
  {TIFA}: Accurate and interpretable text-to-image faithfulness evaluation with
  question answering. In: ICCV (2023)

\bibitem{huang2025comfygpt}
Huang, O., Ma, Y., Zhao, Z., Wu, M., Ji, J., Zhang, R., Hu, Z., Sun, X., Ji,
  R.: {ComfyGPT}: A self-optimizing multi-agent system for comprehensive
  {ComfyUI} workflow generation. arXiv preprint arXiv:2503.17671  (2025)

\bibitem{huang2024smartedit}
Huang, Y., Xie, L., Wang, X., Yuan, Z., Cun, X., Ge, Y., Zhou, J., Dong, C.,
  Huang, R., Zhang, R., et~al.: {SmartEdit}: Exploring complex
  instruction-based image editing with multimodal large language models. In:
  CVPR (2024)

\bibitem{jaech2024openai}
Jaech, A., Kalai, A., Lerer, A., Richardson, A., El-Kishky, A., Low, A.,
  Helyar, A., Madry, A., Beutel, A., Carney, A., et~al.: {OpenAI} o1 {System
  Card}. arXiv preprint arXiv:2412.16720  (2024)

\bibitem{kamath2025geneval}
Kamath, A., Chang, K.W., Krishna, R., Zettlemoyer, L., Hu, Y., Ghazvininejad,
  M.: {GenEval} 2: Addressing benchmark drift in text-to-image evaluation.
  arXiv preprint arXiv:2512.16853  (2025)

\bibitem{kampf2025experiment}
Kampf, K., Brichtova, N.: Experiment with gemini 2.0 flash native image
  generation. Google Developers Blog  (2025)

\bibitem{kong2024hunyuanvideo}
Kong, W., Tian, Q., Zhang, Z., Min, R., Dai, Z., Zhou, J., Xiong, J., Li, X.,
  Wu, B., Zhang, J., et~al.: {HunyuanVideo}: A systematic framework for large
  video generative models. arXiv preprint arXiv:2412.03603  (2024)

\bibitem{kou2024orthus}
Kou, S., Jin, J., Liu, Z., Liu, C., Ma, Y., Jia, J., Chen, Q., Jiang, P., Deng,
  Z.: Orthus: Autoregressive interleaved image-text generation with
  modality-specific heads. arXiv preprint arXiv:2412.00127  (2024)

\bibitem{labs2025flux1kontextflowmatching}
Labs, B.F., Batifol, S., Blattmann, A., Boesel, F., Consul, S., Diagne, C.,
  Dockhorn, T., English, J., English, Z., Esser, P., et~al.: {FLUX.1 Kontext}:
  Flow matching for in-context image generation and editing in latent space.
  arXiv preprint arXiv:2506.15742  (2025)

\bibitem{lewis2020retrieval}
Lewis, P., Perez, E., Piktus, A., Petroni, F., Karpukhin, V., Goyal, N.,
  K{\"u}ttler, H., Lewis, M., Yih, W.t., Rockt{\"a}schel, T., et~al.:
  Retrieval-augmented generation for knowledge-intensive {NLP} tasks. In:
  NeurIPS (2020)

\bibitem{li2024playground}
Li, D., Kamko, A., Akhgari, E., Sabet, A., Xu, L., Doshi, S.: {Playground}
  v2.5: Three insights towards enhancing aesthetic quality in text-to-image
  generation. arXiv preprint arXiv:2402.17245  (2024)

\bibitem{li2025search}
Li, X., Dong, G., Jin, J., Zhang, Y., Zhou, Y., Zhu, Y., Zhang, P., Dou, Z.:
  {Search-o1}: Agentic search-enhanced large reasoning models. In: EMNLP (2025)

\bibitem{li2025torl}
Li, X., Zou, H., Liu, P.: {ToRL}: Scaling tool-integrated {RL}. arXiv preprint
  arXiv:2503.23383  (2025)

\bibitem{li2025perception}
Li, Y., Liu, Z., Li, Z., Zhang, X., Xu, Z., Chen, X., Shi, H., Jiang, S., Wang,
  X., Wang, J., et~al.: Perception, reason, think, and plan: A survey on large
  multimodal reasoning models. arXiv preprint arXiv:2505.04921  (2025)

\bibitem{liu2025advances}
Liu, B., Li, X., Zhang, J., Wang, J., He, T., Hong, S., Liu, H., Zhang, S.,
  Song, K., Zhu, K., et~al.: Advances and challenges in foundation agents: From
  brain-inspired intelligence to evolutionary, collaborative, and safe systems.
  arXiv preprint arXiv:2504.01990  (2025)

\bibitem{liu2025world}
Liu, H., Yan, W., Zaharia, M., Abbeel, P.: World model on million-length video
  and language with blockwise ringattention. In: ICLR (2025)

\bibitem{liu2025step1x}
Liu, S., Han, Y., Xing, P., Yin, F., Wang, R., Cheng, W., Liao, J., Wang, Y.,
  Fu, H., Han, C., et~al.: {Step1X-Edit}: A practical framework for general
  image editing. arXiv preprint arXiv:2504.17761  (2025)

\bibitem{liu2024agentbench}
Liu, X., Yu, H., Zhang, H., Xu, Y., Lei, X., Lai, H., Gu, Y., Ding, H., Men,
  K., Yang, K., et~al.: {AgentBench}: Evaluating {LLM}s as agents. In: ICLR
  (2024)

\bibitem{liu2025breaking}
Liu, Y., Cao, J., Li, Z., He, R., Tan, T.: Breaking mental set to improve
  reasoning through diverse multi-agent debate. In: ICLR (2025)

\bibitem{ma2025janusflow}
Ma, Y., Liu, X., Chen, X., Liu, W., Wu, C., Wu, Z., Pan, Z., Xie, Z., Zhang,
  H., Yu, X., et~al.: {JanusFlow}: Harmonizing autoregression and rectified
  flow for unified multimodal understanding and generation. In: CVPR (2025)

\bibitem{niu2025flow}
Niu, B., Song, Y., Lian, K., Shen, Y., Yao, Y., Zhang, K., Liu, T.: Flow:
  Modularized agentic workflow automation. arXiv preprint arXiv:2501.07834
  (2025)

\bibitem{niu2025wise}
Niu, Y., Ning, M., Zheng, M., Jin, W., Lin, B., Jin, P., Liao, J., Feng, C.,
  Meng, F., Ning, K., et~al.: {WISE}: A world knowledge-informed semantic
  evaluation for text-to-image generation. arXiv preprint arXiv:2503.07265
  (2025)

\bibitem{gpt_image_1}
OpenAI: gpt-image-1 (2025),
  \url{https://platform.openai.com/docs/models/gpt-image-1}

\bibitem{openai2025deep_research}
{OpenAI}: Introducing {Deep Research} (2025),
  \url{https://openai.com/index/introducing-deep-research/}

\bibitem{openai2025o3o4}
{OpenAI}: Introducing {OpenAI} o3 and o4-mini (2025),
  \url{https://openai.com/index/introducing-o3-and-o4-mini/}

\bibitem{patil2024gorilla}
Patil, S.G., Zhang, T., Wang, X., Gonzalez, J.E.: {Gorilla}: Large language
  model connected with massive {API}s. In: NeurIPS (2024)

\bibitem{podell2024sdxl}
Podell, D., English, Z., Lacey, K., Blattmann, A., Dockhorn, T., M{\"u}ller,
  J., Penna, J., Rombach, R.: {SDXL}: Improving latent diffusion models for
  high-resolution image synthesis. In: ICLR (2024)

\bibitem{qian2024chatdev}
Qian, C., Liu, W., Liu, H., Chen, N., Dang, Y., Li, J., Yang, C., Chen, W., Su,
  Y., Cong, X., et~al.: {ChatDev}: Communicative agents for software
  development. In: ACL (2024)

\bibitem{qiao2025benchmarking}
Qiao, S., Fang, R., Qiu, Z., Wang, X., Zhang, N., Jiang, Y., Xie, P., Huang,
  F., Chen, H.: Benchmarking agentic workflow generation. In: ICLR (2025)

\bibitem{qiu2023controlling}
Qiu, Z., Liu, W., Feng, H., Xue, Y., Feng, Y., Liu, Z., Zhang, D., Weller, A.,
  Sch{\"o}lkopf, B.: Controlling text-to-image diffusion by orthogonal
  finetuning. In: NeurIPS (2023)

\bibitem{ramesh2022hierarchical}
Ramesh, A., Dhariwal, P., Nichol, A., Chu, C., Chen, M.: Hierarchical
  text-conditional image generation with {CLIP} latents. arXiv preprint
  arXiv:2204.06125  (2022)

\bibitem{rombach2022high}
Rombach, R., Blattmann, A., Lorenz, D., Esser, P., Ommer, B.: High-resolution
  image synthesis with latent diffusion models. In: CVPR (2022)

\bibitem{schick2023toolformer}
Schick, T., Dwivedi-Yu, J., Dess{\`\i}, R., Raileanu, R., Lomeli, M., Hambro,
  E., Zettlemoyer, L., Cancedda, N., Scialom, T.: Toolformer: Language models
  can teach themselves to use tools. In: NeurIPS (2023)

\bibitem{shang2025agentsquare}
Shang, Y., Li, Y., Zhao, K., Ma, L., Liu, J., Xu, F., Li, Y.: {AgentSquare}:
  Automatic {LLM} agent search in modular design space. In: ICLR (2025)

\bibitem{shinn2023reflexion}
Shinn, N., Cassano, F., Gopinath, A., Narasimhan, K., Yao, S.: Reflexion:
  Language agents with verbal reinforcement learning. In: NeurIPS (2023)

\bibitem{autogpt}
{Significant Gravitas}: {AutoGPT} (2023),
  \url{https://github.com/Significant-Gravitas/AutoGPT}

\bibitem{sun2024autoregressive}
Sun, P., Jiang, Y., Chen, S., Zhang, S., Peng, B., Luo, P., Yuan, Z.:
  Autoregressive model beats diffusion: {Llama} for scalable image generation.
  arXiv preprint arXiv:2406.06525  (2024)

\bibitem{sun2024generative}
Sun, Q., Cui, Y., Zhang, X., Zhang, F., Yu, Q., Wang, Y., Rao, Y., Liu, J.,
  Huang, T., Wang, X.: Generative multimodal models are in-context learners.
  In: CVPR (2024)

\bibitem{team2024chameleon}
Team, C.: Chameleon: Mixed-modal early-fusion foundation models. arXiv preprint
  arXiv:2405.09818  (2024)

\bibitem{wang2025mixture}
Wang, J., Wang, J., Athiwaratkun, B., Zhang, C., Zou, J.Y.: Mixture-of-agents
  enhances large language model capabilities. In: ICLR (2025)

\bibitem{wang2024rethinking}
Wang, Q., Wang, Z., Su, Y., Tong, H., Song, Y.: Rethinking the bounds of {LLM}
  reasoning: Are multi-agent discussions the key? In: ACL (2024)

\bibitem{wang2024emu3}
Wang, X., Zhang, X., Luo, Z., Sun, Q., Cui, Y., Wang, J., Zhang, F., Wang, Y.,
  Li, Z., Yu, Q., et~al.: {Emu3}: Next-token prediction is all you need. arXiv
  preprint arXiv:2409.18869  (2024)

\bibitem{wang2022self}
Wang, X., Wei, J., Schuurmans, D., Le, Q., Chi, E., Narang, S., Chowdhery, A.,
  Zhou, D.: Self-consistency improves chain of thought reasoning in language
  models. arXiv preprint arXiv:2203.11171  (2022)

\bibitem{wei2022chain}
Wei, J., Wang, X., Schuurmans, D., Bosma, M., Xia, F., Chi, E., Le, Q.V., Zhou,
  D., et~al.: Chain-of-thought prompting elicits reasoning in large language
  models. In: NeurIPS (2022)

\bibitem{de2024system}
de~Winter, J.C., Dodou, D., Eisma, Y.B.: System 2 thinking in openai’s
  o1-preview model: Near-perfect performance on a mathematics exam. Computers
  \textbf{13}(11), ~278 (2024)

\bibitem{wu2025qwen}
Wu, C., Li, J., Zhou, J., Lin, J., Gao, K., Yan, K., Yin, S.m., Bai, S., Xu,
  X., Chen, Y., et~al.: {Qwen-Image} technical report. arXiv preprint
  arXiv:2508.02324  (2025)

\bibitem{wu2025janus}
Wu, C., Chen, X., Wu, Z., Ma, Y., Liu, X., Pan, Z., Liu, W., Xie, Z., Yu, X.,
  Ruan, C., et~al.: Janus: Decoupling visual encoding for unified multimodal
  understanding and generation. In: CVPR (2025)

\bibitem{wu2023autogen}
Wu, Q., Bansal, G., Zhang, J., Wu, Y., Li, B., Zhu, E., Jiang, L., Zhang, X.,
  Zhang, S., Liu, J., et~al.: {AutoGen}: Enabling next-gen {LLM} applications
  via multi-agent conversation. arXiv preprint arXiv:2308.08155  (2023)

\bibitem{wu2025vila}
Wu, Y., Zhang, Z., Chen, J., Tang, H., Li, D., Fang, Y., Zhu, L., Xie, E., Yin,
  H., Yi, L., et~al.: {VILA-U}: a unified foundation model integrating visual
  understanding and generation. In: ICLR (2025)

\bibitem{wu2026kris}
Wu, Y., Li, Z., Hu, X., Ye, X., Zeng, X., Yu, G., Zhu, W., Schiele, B., Yang,
  M.H., Yang, X.: {KRIS-Bench}: Benchmarking next-level intelligent image
  editing models. In: NeurIPS (2026)

\bibitem{xiao2025omnigen}
Xiao, S., Wang, Y., Zhou, J., Yuan, H., Xing, X., Yan, R., Li, C., Wang, S.,
  Huang, T., Liu, Z.: {OmniGen}: Unified image generation. In: CVPR (2025)

\bibitem{xiao2025verbalized}
Xiao, T.Z., Bamler, R., Sch{\"o}lkopf, B., Liu, W.: Verbalized machine
  learning: Revisiting machine learning with language models. Transactions on
  Machine Learning Research  (2025)

\bibitem{xie2025show}
Xie, J., Mao, W., Bai, Z., Zhang, D.J., Wang, W., Lin, K.Q., Gu, Y., Chen, Z.,
  Yang, Z., Shou, M.Z.: {Show-o}: One single transformer to unify multimodal
  understanding and generation. In: ICLR (2025)

\bibitem{xie2023openagents}
Xie, T., Zhou, F., Cheng, Z., Shi, P., Weng, L., Liu, Y., Hua, T.J., Zhao, J.,
  Liu, Q., Liu, C., et~al.: {OpenAgents}: An open platform for language agents
  in the wild. arXiv preprint arXiv:2310.10634  (2023)

\bibitem{xu2023towards}
Xu, Z., Shi, S., Hu, B., Yu, J., Li, D., Zhang, M., Wu, Y.: Towards reasoning
  in large language models via multi-agent peer review collaboration. arXiv
  preprint arXiv:2311.08152  (2023)

\bibitem{xu2025comfyui}
Xu, Z., Yangxue, Y., Wang, Y., Hu, Q., Wu, Z., Hu, B., Wang, L., Luo, W.,
  Zhang, K.: {ComfyUI-Copilot}: An intelligent assistant for automated workflow
  development. In: ACL (2025)

\bibitem{xue2025comfybench}
Xue, X., Lu, Z., Huang, D., Wang, Z., Ouyang, W., Bai, L.: {ComfyBench}:
  Benchmarking {LLM}-based agents in {ComfyUI} for autonomously designing
  collaborative {AI} systems. In: CVPR (2025)

\bibitem{yang2025qwen3}
Yang, A., Li, A., Yang, B., Zhang, B., Hui, B., Zheng, B., Yu, B., Gao, C.,
  Huang, C., Lv, C., et~al.: {Qwen3} technical report. arXiv preprint
  arXiv:2505.09388  (2025)

\bibitem{yao2022react}
Yao, S., Zhao, J., Yu, D., Du, N., Shafran, I., Narasimhan, K., Cao, Y.:
  {ReAct}: Synergizing reasoning and acting in language models. arXiv preprint
  arXiv:2210.03629  (2022)

\bibitem{ye2023ip}
Ye, H., Zhang, J., Liu, S., Han, X., Yang, W.: {IP-Adapter}: Text compatible
  image prompt adapter for text-to-image diffusion models. arXiv preprint
  arXiv:2308.06721  (2023)

\bibitem{yu2025anyedit}
Yu, Q., Chow, W., Yue, Z., Pan, K., Wu, Y., Wan, X., Li, J., Tang, S., Zhang,
  H., Zhuang, Y.: {AnyEdit}: Mastering unified high-quality image editing for
  any idea. In: CVPR (2025)

\bibitem{yu2025generating}
Yu, Z., Yuan, Y., Xiao, T.Z., Xia, F.F., Fu, J., Zhang, G., Lin, G., Liu, W.:
  Generating symbolic world models via test-time scaling of large language
  models. Transactions on Machine Learning Research  (2025)

\bibitem{yuksekgonul2024textgrad}
Yuksekgonul, M., Bianchi, F., Boen, J., Liu, S., Huang, Z., Guestrin, C., Zou,
  J.: Textgrad: Automatic "differentiation" via text. arXiv preprint
  arXiv:2406.07496  (2024)

\bibitem{zhang2025aflow}
Zhang, J., Xiang, J., Yu, Z., Teng, F., Chen, X., Chen, J., Zhuge, M., Cheng,
  X., Hong, S., Wang, J., et~al.: {AFlow}: Automating agentic workflow
  generation. In: ICLR (2025)

\bibitem{zhang2023magicbrush}
Zhang, K., Mo, L., Chen, W., Sun, H., Su, Y.: {MagicBrush}: A manually
  annotated dataset for instruction-guided image editing. In: NeurIPS (2023)

\bibitem{zhang2023adding}
Zhang, L., Rao, A., Agrawala, M.: Adding conditional control to text-to-image
  diffusion models. In: ICCV (2023)

\bibitem{zhao2024marco}
Zhao, Y., Yin, H., Zeng, B., Wang, H., Shi, T., Lyu, C., Wang, L., Luo, W.,
  Zhang, K.: {Marco-o1}: Towards open reasoning models for open-ended
  solutions. arXiv preprint arXiv:2411.14405  (2024)

\bibitem{zheng2025deepresearcher}
Zheng, Y., Fu, D., Hu, X., Cai, X., Ye, L., Lu, P., Liu, P.: {DeepResearcher}:
  Scaling deep research via reinforcement learning in real-world environments.
  In: EMNLP (2025)

\end{thebibliography}
